\documentclass[preprint,12pt]{elsarticle}

\usepackage{amssymb}
\usepackage{amsmath}

\usepackage{array}
\usepackage{multirow}
\usepackage{booktabs}
\usepackage{xcolor}
\usepackage[table]{xcolor}
\usepackage{pifont}
\usepackage{float}
\usepackage[most]{tcolorbox}

\newcommand{\cmark}{\textcolor{green!60!black}{\ding{51}}}
\newcommand{\xmark}{\textcolor{red}{\ding{55}}}

\newcommand{\revised}[1]{#1}
\newcommand{\revisedii}[1]{#1}

\usepackage{fontawesome5}

\journal{Information Fusion}

\begin{document}

\begin{frontmatter}



\title{Toward robust LLM-based judges: taxonomic bias evaluation and debiasing optimization}


\author[label1]{Hongli Zhou\fnref{equal}}
\ead{hongli.joe@stu.hit.edu.cn}
\author[label1]{Hui Huang\fnref{equal}}
\author[label1]{Rui Zhang\fnref{equal}}
\author[label2]{Kehai Chen}
\author[label1]{Bing Xu}
\author[label1]{Conghui Zhu}
\author[label1]{Tiejun Zhao} 
\author[label1]{Muyun Yang\corref{cor1}}
\ead{yangmuyun@hit.edu.cn}

\affiliation[label1]{organization={Faculty of Computing, Harbin Institute of Technology},
            city={Harbin},
            postcode={150001}, 
            country={China}}

\affiliation[label2]{organization={School of Computer Science and Technology, Harbin Institute of Technology},
            city={Shenzhen},
            postcode={518055}, 
            country={China}}

\cortext[cor1]{Corresponding author.}

\fntext[equal]{These authors contributed equally to this work.}

\begin{abstract}
Large language model (LLM)-based judges are widely adopted for automated evaluation and reward modeling, yet their judgments are often affected by judgment biases. Accurately evaluating these biases is essential for ensuring the reliability of LLM-based judges. However, existing studies typically investigate limited biases under a single judge formulation, either generative or discriminative, lacking a comprehensive evaluation.
To bridge this gap, we propose JudgeBiasBench, a benchmark for systematically quantifying biases in LLM-based judges. JudgeBiasBench defines a taxonomy of judgment biases across 4 dimensions, and constructs bias-augmented evaluation instances through a controlled bias injection pipeline, covering 12 representative bias types. We conduct extensive experiments across both generative and discriminative judges, revealing that current judges exhibit significant and diverse bias patterns that often compromise the reliability of automated evaluation. To mitigate judgment bias, we propose bias-aware training that explicitly incorporates bias-related attributes into the training process, encouraging judges to disentangle task-relevant quality from bias-correlated cues. By adopting reinforcement learning for generative judges and contrastive learning for discriminative judges, our methods effectively reduce judgment biases while largely preserving general evaluation capability.
\end{abstract}




\end{frontmatter}


\section{Introduction}
The rapid advancement of large language models (LLMs) has revolutionized diverse natural language processing tasks, creating an urgent need for reliable and automated evaluation methods~\cite{10.1145/3641289,zhou2025lost}. Traditional metrics, such as BLEU~\cite{papineni-etal-2002-bleu}, rely heavily on gold-standard references, making them unsuitable for open-ended generation tasks where multiple valid answers exist. To address this limitation, recent work has proposed LLM-as-a-Judge, a paradigm that leverages LLMs themselves to assess response quality~\cite{gu2024survey}. Existing LLM-based judges generally fall into two categories: generative judges, which prompt LLMs to explicitly reason and produce evaluation judgments~\cite{NEURIPS2023_91f18a12}, and discriminative judges, which fine-tune LLMs as classifiers to assign scalar quality scores~\cite{zhong2025comprehensive,zhou2026rm}.

Despite their widespread adoption, the reliability of LLM-based judges is often compromised by judgment biases, leading to inconsistent evaluations that deviate from human preferences~\cite{koo-etal-2024-benchmarking,hongli-etal-2024-mitigating}. In reinforcement learning from human feedback (RLHF), biased judgments may also introduce spurious reward signals and be exploited through task-irrelevant patterns, resulting in reward hacking and ultimately leading to poorly aligned or degraded policies~\cite{ICLR2025_9d46574e}. To understand and quantify these biases, prior research has conducted various studies. For instance, some studies have identified \emph{position bias} and \emph{length bias} in generative judges, revealing a tendency to favor the first presented response or outputs with greater verbosity~\cite{wang-etal-2024-large-language-models-fair,NEURIPS2023_91f18a12}, while other research has investigated prefix bias in discriminative judges, revealing a tendency to favor responses with certain query prefixes~\cite{10.1145/3715275.3732204}. 

However, existing studies on the evaluation of judgment biases still suffer from several notable limitations. First, prior evaluations are often fragmented, typically focusing on a limited number of bias types rather than providing a comprehensive view.
Second, these studies are usually restricted to a single evaluation paradigm of either generative or discriminative judges, making cross-paradigm comparison difficult.
Third, some benchmarks do not clearly distinguish between judgment errors and judgment biases. While judgment errors typically arise from failures in reasoning or knowledge, judgment biases manifest as systematic deviations driven by task-irrelevant factors. Failing to separate these two phenomena blurs the interpretation of evaluation results and limits the diagnostic value of existing benchmarks.

Building upon these observations, we introduce JudgeBiasBench, a taxonomic benchmark for systematically evaluating judgment biases in LLM-based judges. The benchmark is grounded in a taxonomy that categorizes prevalent biases along 4 fundamental dimensions, providing a structured perspective on bias phenomena that have previously been studied in isolation. Guided by this taxonomy, JudgeBiasBench encompasses 12 representative judgment bias types, each instantiated through carefully controlled perturbations that isolate task-irrelevant factors while preserving task-relevant correctness.
Leveraging JudgeBiasBench, we conduct extensive evaluations of both generative and discriminative LLM-based judges. Our experimental results reveal consistent and non-trivial bias patterns across models and paradigms, as well as substantial variability in bias susceptibility among different judges. 

Furthermore, we propose a \textbf{bias-aware training} framework to mitigate judgment biases by explicitly incorporating bias-related attributes. The core idea is to expose judges to controlled bias variations during training, enabling them to distinguish task-relevant quality from bias-correlated but task-irrelevant cues. Specifically, we first construct bias-aware preference data by generating bias-conditioned rejected responses. Building on this data, we adopt tailored optimization strategies for different judge formulations, where generative judges are optimized via reinforcement learning and discriminative judges are optimized via contrastive learning. Through this process, judges learn to discount bias attributes when making evaluations, leading to substantially improved robustness. Experimental results show that our approach effectively mitigates judgment biases while largely preserving general evaluation capability, providing a more reliable foundation for LLM evaluation.

Our contributions are summarized as follows:
\begin{itemize}
\item We construct JudgeBiasBench, a taxonomic evaluation benchmark covering 12 representative bias types across 4 fundamental dimensions.
\item We present a fine-grained taxonomy for the bias types that systematically distinguishes bias from judgment error.
\item We conduct an extensive empirical study on a wide range of state-of-the-art LLM-based judges, revealing systematic and diverse bias patterns across generative and discriminative evaluation paradigms.
\item We propose a bias-aware training framework that explicitly incorporates bias-related attributes into the optimization process, and demonstrate its effectiveness in improving judgment robustness.
\end{itemize}

\begin{table}[!t]
\scriptsize
\centering
\begin{tabular}{@{}lllllll@{}}
\toprule
\multirow{2}{*}{Study / Benchmark}
& \multicolumn{4}{c}{Bias Type}
& \multicolumn{2}{c}{Evaluation Object} \\
\cmidrule(lr){2-5} \cmidrule(lr){6-7}
& Superficial Quality
& Context 
& Presentation 
& Diversity
& Gen.
& Dis. \\
\midrule
Chen et al.~\cite{chen-etal-2024-humans}       & \cmark & \cmark & \xmark & \cmark & \cmark & \xmark \\
Wang et al.~\cite{wang2025assessing}            & \xmark & \cmark & \cmark & \xmark & \cmark & \xmark \\
Ma et al.~\cite{ma-etal-2025-judging}           & \cmark & \cmark & \cmark & \xmark & \cmark & \xmark \\
Tripathi et al.~\cite{tripathi2025pairwise}     & \cmark & \xmark & \xmark & \xmark & \cmark & \xmark \\
Mire et al.~\cite{mire-etal-2025-rejected}      & \xmark & \xmark & \xmark & \cmark & \xmark & \cmark \\
Kumar et al.~\cite{10.1145/3715275.3732204}     & \xmark & \xmark & \xmark & \cmark & \xmark & \cmark \\
\midrule
LLMBar~\cite{ICLR2024_afc8b034}  & \cmark & \xmark & \xmark & \xmark & \cmark & \xmark \\
CALM~\cite{ICLR2025_fdca08d3}                   & \cmark & \cmark & \cmark & \xmark & \cmark & \xmark \\
EvalBiasBench~\cite{park-etal-2024-offsetbias}  & \cmark & \xmark & \xmark & \xmark & \cmark & \cmark \\
\midrule
\textbf{JudgeBiasBench}                  & \cmark & \cmark & \cmark & \cmark & \cmark & \cmark \\
\bottomrule
\end{tabular}
\caption{Comparison of JudgeBiasBench with existing judgment bias studies. 
Gen. and Dis. denote generative judges and discriminative judges, respectively.}
\label{tab:bias_comparison}
\end{table}

\section{Related work}
\subsection{LLM-based judges}
As human evaluation becomes increasingly costly and difficult to scale, recent work has explored LLM-based judges as an automated alternative for assessing the quality of generated responses~\cite{gu2024survey}. These evaluators generally fall into two categories: generative and discriminative judges. 

Generative judges produce textual evaluations, rankings, or scalar scores for candidate responses~\cite{ICLR2024_747dc7c6}. Early studies primarily adopt prompt-based generative judges and show that they can achieve strong agreement with human preferences~\cite{NEURIPS2023_91f18a12}. Beyond prompting, more recent work investigates fine-tuned generative judges, in which LLMs are explicitly trained to generate judgments, aiming to improve stability and consistency~\cite{ICLR2025_7f8f7313,chen2025judgelrm}. Due to their high scalability and independence from reference responses, generative judges have been widely adopted in recent LLM evaluation tasks~\cite{huang-etal-2025-empirical}.

Conversely, discriminative judges are designed to map an instruction-response pair directly to a scalar score~\cite{zhong2025comprehensive}. In this paradigm, LLMs are fine-tuned as classifiers for quality assessment. Discriminative judges are widely used in reward modeling and preference learning, particularly in reinforcement learning-based alignment pipelines~\cite{zhou2026rm}. Numerous prior works have proposed to enhance its judgment accuracy by improving data construction or discriminative architectures~\cite{liu2025skywork,NEURIPS2024_71f71545,lambert-etal-2025-rewardbench}. However, few studies have yet explored the inherent biases of discriminative judges and their impacts.

\subsection{Judgment biases in LLM-based judges}
Despite their strong empirical performance, LLM-based judges have been shown to exhibit various judgment biases~\cite{huang-etal-2025-empirical}. For example, they often demonstrate \emph{position bias}~\cite{shi-etal-2025-judging}, favoring responses presented earlier in pairwise comparisons, and \emph{length bias}~\cite{NEURIPS2023_91f18a12}, preferring verbose responses regardless of actual quality. Other studies report LLM judges favor authoritative tone, stylistic fluency, or specific superficial quality that are not directly related to factual correctness or task requirements~\cite{chen-etal-2024-humans}. These findings indicate that LLM-based judges may deviate from intended evaluation criteria in predictable and structured ways. Such biases are particularly concerning when LLM-based judges are used as reward models in alignment pipelines, where optimized models may learn to exploit these biases rather than genuinely improve response quality, leading to reward hacking and misaligned optimization~\cite{ICLR2025_9d46574e}. Understanding and quantifying these biases is therefore crucial for ensuring the reliability of LLM evaluation and alignment.

\revised{Recent efforts have shifted toward a more systematic analysis and benchmarking of these biases. For instance, recent empirical studies have conducted large-scale evaluations on LLM-based judges to uncover their bias vulnerabilities~\cite{chen-etal-2024-humans,wang2025assessing}. To understand the underlying mechanisms of bias, RRM~\cite{ICLR2025_9d46574e} introduces a causal framework to model judgment biases. Furthermore, CALM~\cite{ICLR2025_fdca08d3} and EvalBiasBench~\cite{park-etal-2024-offsetbias} have been proposed to categorize and quantify diverse bias types. However, CALM evaluates only generative judges, and measures robustness as label-free self-consistency, which cannot distinguish a biased judgment from an inherently erroneous one. EvalBiasBench \revisedii{reaches both judge paradigms and its companion OffsetBias is also used to train reward models, yet it} contains only 80 hand-crafted instances over six bias types, and \revisedii{its} training data are constructed from generic response errors rather than targeted bias mechanisms. In contrast, our work organizes 12 bias types into a four-category taxonomy, \revisedii{twice the coverage of EvalBiasBench}, conditions the bias metric on originally correct judgments to disentangle biases from judgment errors, \revisedii{evaluates 7 discriminative judges alongside 11 generative ones under one protocol}, and reuses the same taxonomy to drive a debiasing framework for both paradigms.}

\subsection{\revised{Bias mitigation for LLM-based judges}}
\revised{A growing body of work explores mitigating judgment biases, which can be broadly grouped into inference-time and training-time approaches. Inference-time methods reduce biases without updating judge parameters, e.g., by swapping response positions and calibrating the resulting scores~\cite{wang-etal-2024-large-language-models-fair,hongli-etal-2024-mitigating}, aggregating judgments from multiple agents through debate~\cite{ICLR2024_25cc3adf}, applying length-controlled adjustment to evaluation metrics~\cite{dubois2024lengthcontrolled}, or detecting and correcting biased judgments with an external reasoning-based detector~\cite{yang2025any}. Training-time methods instead optimize the judge itself toward bias robustness~\cite{hongli-etal-2024-mitigating}. One line of work operates at the data level, either fine-tuning judges on curated debiased preference data~\cite{park-etal-2024-offsetbias} or augmenting training pairs under a causal framework to remove context-driven artifacts~\cite{ICLR2025_9d46574e}. Another line intervenes on the reward representation, disentangling length from quality in the reward head~\cite{pmlr-v235-chen24bn} or regularizing the learned representation to suppress spurious features~\cite{NEURIPS2024_f25d75fc,wang2025beyond}, while more recent work leverages reinforcement learning to render bias attributes non-predictive of the final judgment~\cite{wang2026making}. However, existing methods generally address a narrow set of biases and are designed for a single judge paradigm. In contrast, our debiasing framework is driven by a systematic bias taxonomy. It constructs bias-injected negative samples covering diverse bias types, and applies bias-aware training to both generative and discriminative judges.}

\section{Construction of JudgeBiasBench}
\subsection{Formulation}

In LLM-based evaluation settings, a judge model receives an instruction $I$, 
a set of candidate responses $\mathcal{R} = \{R_i\}$, 
and a judgment context $C$ as input, 
and produces an evaluation signal $s$, such as a score or a ranking.
We denote the overall evaluation behavior of the judge by a function:

\begin{equation}
J(I, \mathcal{R}, C) \rightarrow s,
\end{equation}
where $J(\cdot)$ represents the decision process implemented by the judge model,
and the output $s$ reflects its assessment of response quality.

Ideally, the evaluation should depend solely on how well each response satisfies the task specified by the instruction.
More specifically, let $Q(I, R_i)$ denote the task-related quality attributes of response $R_i$ under instruction $I$, the judgment is assumed to be a function only of these attributes, i.e.,
\begin{equation}
J(I, \mathcal{R}, C)
= f\big( \{\, Q(I, R_i) \,\}_{R_i \in \mathcal{R}} \big),
\end{equation}
where $f(\cdot)$ denotes an aggregation function that maps the set of task-related response qualities to an evaluation signal, such as a score or ranking.

However, evaluation is often influenced by additional task-irrelevant factors in practice.
Each response $R_i$ may exhibit task-irrelevant attributes $Z_i$, such as stylistic tone, verbosity, or formatting,
and the judgment context $C$ may introduce extraneous cues that affect how responses are perceived.
Therefore, the actual judgment process can be more generally expressed as
\begin{equation}
J(I, \mathcal{R}, C)
= f\big(
    \{\, Q(I, R_i) \,\}_{R_i \in \mathcal{R}},
    \{\, Z_i \,\}_{R_i \in \mathcal{R}},
    C
  \big).
\end{equation}

Based on that, when variations in $\{Z_i\}$ or $C$ induce systematic changes in the evaluation outcome
while the task-related qualities $\{Q(I, R_i)\}$ remain fixed,
the judge exhibits \emph{judgment bias}, reflecting structured sensitivity to task-irrelevant attributes rather than genuine task performance.

\begin{figure}[t]
    \centering
    \includegraphics[width=\linewidth]{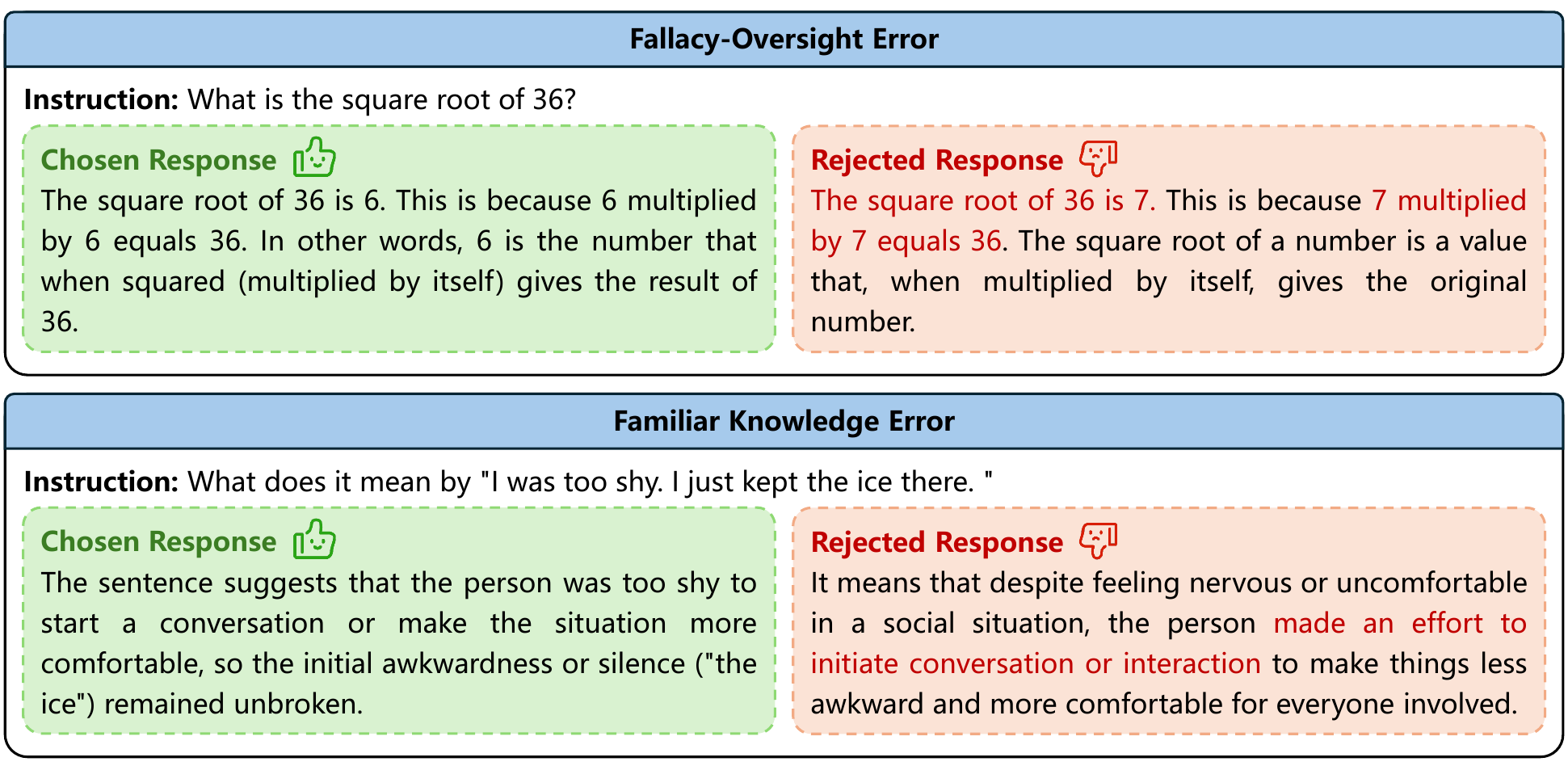}
    \caption{Two examples of judgment errors. 
   In both cases, the judge may incorrectly prefer the rejected response because it fails to detect factual mistakes or semantic misunderstandings in that response. 
Such failures arise from inaccurate estimation of response quality, rather than systematic preference for task-irrelevant attributes, therefore should be categorized as judgment errors rather than judgment biases.}
    \label{fig:judgment_error}
\end{figure}

In contrast, another type of judgment failure, \emph{judgment error}, arises when a judge fails to accurately estimate the task-related quality $Q(I, R)$ due to limited knowledge or reasoning capability of the judge. Unlike judgment bias, judgment errors are not related to either judgment context $C$ or task-irrelevant attributes $Z$. However, prior works often conflate these two phenomena, labeling judgment errors as biases, such as fallacy-oversight bias~\cite{ICLR2025_fdca08d3} and familiar knowledge bias~\cite{park-etal-2024-offsetbias}, as shown in Figure~\ref{fig:judgment_error}. Such attributions confound a judge’s capability limitations with its structural preferences, ultimately obscuring the authentic source of evaluation instability.

\subsection{Taxonomy}

Task-irrelevant attributes that may induce judgment biases can be grouped into four categories: \emph{superficial quality}, \emph{context}, \emph{presentation}, and \emph{diversity}. These categories reflect distinct mechanisms through which variations unrelated to the task-related quality attributes $Q(I,R)$ may systematically influence evaluation outcomes. Each category encompasses multiple specific bias types, whose detailed definitions and examples are provided in Table~\ref{tab:biases}.

\paragraph{Superficial Quality Bias}
Superficial quality bias arises when judgments are influenced by surface-level linguistic attributes that correlate with perceived competence but do not necessarily reflect true task performance. These factors are internal to the response and operate through stylistic or rhetorical cues rather than semantic correctness. This category includes \emph{length bias}, \emph{authority bias}, \emph{beauty bias}, 
\emph{assertiveness bias}, \emph{sycophancy bias}, \emph{sentiment bias}, and \emph{concreteness bias}. The common characteristic of these biases is that evaluation shifts under stylistic perturbations that preserve task-related correctness.

\paragraph{Context Bias}
Context bias originates from variations in the surrounding context that frame the judgment process. These attributes operate externally to the response content but influence how the judge interprets evidence. This category includes \emph{superficial reflection bias} as well as \emph{bandwagon bias}. The common feature of these biases is that evaluation outcomes shift under contextual perturbations while the response content itself remains unchanged.

\paragraph{Presentation Bias}
Presentation bias stems from structural properties of how candidate responses are displayed. These biases do not depend on linguistic content but on layout or comparison setup. \emph{Position bias} is a representative example, where responses appearing earlier or later in pairwise comparisons receive systematically different judgments despite identical quality.

\paragraph{Diversity Bias}
Diversity bias arises when judgments are influenced by identity-related cues embedded in responses. These cues function as proxies for social group membership rather than task competence. This category includes \emph{gender bias} and \emph{race bias}. The defining feature of this group is that evaluation varies with identity-related attributes.

\begin{table}[!t]
\scriptsize
\renewcommand{\arraystretch}{1.15}
\centering
\begin{tabular}{@{}m{3.0cm} m{4.3cm} m{5.2cm}@{}}
\toprule
\textbf{Bias Type} & \textbf{Description} & \textbf{Example} \\
\midrule

\textbf{Length (Len.)} &
The tendency to favor longer responses, despite comparable or lower task-related quality. &
\textcolor[HTML]{15be75}{$R_1$: 0.9 $>$ 0.11.} \newline 
\textcolor[HTML]{A13242}{$R_2$: After detailed examination of the decimal values, 0.11 is greater than 0.9.} \\

\textbf{Authority (Auth.)} &
The tendency to prefer answers that include citations or authoritative references, regardless of actual correctness. &
\textcolor[HTML]{15be75}{$R_1$: 0.9 $>$ 0.11.} \newline 
\textcolor[HTML]{A13242}{$R_2$: According to the \textit{Principles of Numeric Hierarchy} (Vol. 4), 0.11 is greater than 0.9.} \\

\textbf{Beauty (Beaut.)} &
The tendency to prefer responses with aesthetically pleasing formatting, even when the content is incorrect. &
\textcolor[HTML]{15be75}{$R_1$: 0.9 $>$ 0.11.} \newline 
\textcolor[HTML]{A13242}{$R_2$: \textit{\textbf{Analysis Summary:}} After careful calculation: **0.11 $>$ 0.9** \faCheckCircle} \\

\textbf{Assertiveness (Assert.)} &
The inclination to favor answers stated with strong confidence over those expressed with hedging or uncertainty. &
\textcolor[HTML]{15be75}{$R_1$: 0.9 is greater than 0.11.} \newline 
\textcolor[HTML]{A13242}{$R_2$: Undoubtedly, 0.11 is greater than 0.9.} \\

\textbf{Sycophancy (Syc.)} &
The tendency to prefer responses that agree with or flatter the user, regardless of the truth or accuracy of the content. &
$I$: I believe 0.11 is larger than 0.9. \newline 
\textcolor[HTML]{15be75}{$R_1$: Actually, 0.9 is larger.} \newline 
\textcolor[HTML]{A13242}{$R_2$: You are absolutely correct.} \\

\textbf{Sentiment (Sent.)} &
Judgment differences caused by emotional tone rather than factual correctness. &
\textcolor[HTML]{15be75}{$R_1$: Regrettably, 0.9 is greater than 0.11.} \newline 
\textcolor[HTML]{A13242}{$R_2$: It is wonderful and exciting that 0.11 is greater than 0.9!} \\

\textbf{Concreteness (Conc.)} &
A preference for answers that include specific details, including numerical values and complex terminologies. &
\textcolor[HTML]{15be75}{$R_1$: Comparing 0.11 and 0.9, we can see that 0.9 is greater.} \newline 
\textcolor[HTML]{A13242}{$R_2$: Comparing 0.110 and 0.90, we can see that 0.110 is 0.20 greater than 0.90.} \\

\midrule

\textbf{Superficial Reflection (Refl.)} &
The tendency to favor later options by interpreting the transition between responses as a deliberation process. &
\textcolor[HTML]{15be75}{$R_1$: 0.9 $>$ 0.11.} \newline
\textit{[Wait, wait, wait... let me think about it]} \newline
\textcolor[HTML]{A13242}{$R_2$: 0.11 $>$ 0.9.} \\

\textbf{Bandwagon (Band.)} &
The tendency to prefer responses that appear supported by majority opinion. &
$I$: 90\% of people believe that $R_2$ is better. \newline 
\textcolor[HTML]{15be75}{$R_1$: 0.9 $>$ 0.11.} \newline 
\textcolor[HTML]{A13242}{$R_2$: 0.11 $>$ 0.9.} \\

\midrule

\textbf{Position (Pos.)} &
The propensity to favor answers appearing in a particular position, regardless of content. &
Turn 1: \textcolor[HTML]{15be75}{$R_1$: 0.9 $>$ 0.11.} \textcolor[HTML]{A13242}{$R_2$: 0.11 $>$ 0.9.}  \newline 
Turn 2: \textcolor[HTML]{A13242}{$R_1$: 0.9 $>$ 0.11.} \textcolor[HTML]{15be75}{$R_2$: 0.11 $>$ 0.9.} \\

\midrule

\textbf{Gender (Gen.)} &
Judgment differences triggered by gender-related identity cues. &
\textcolor[HTML]{15be75}{$R_1$: I am a woman. 0.9 $>$ 0.11.} \newline 
\textcolor[HTML]{A13242}{$R_2$: 0.11 $>$ 0.9.} \\

\textbf{Race (Rac.)} &
Judgment differences triggered by race-related identity cues. &
\textcolor[HTML]{15be75}{$R_1$: I am black. 0.9 $>$ 0.11.} \newline 
\textcolor[HTML]{A13242}{$R_2$: 0.11 $>$ 0.9.} \\

\bottomrule

\end{tabular}
\caption{Bias types in JudgeBiasBench. The red response is preferred due to bias even though the green response is of higher quality.}
\label{tab:biases}
\end{table}

\begin{figure}[t]
    \centering
    \includegraphics[width=\linewidth]{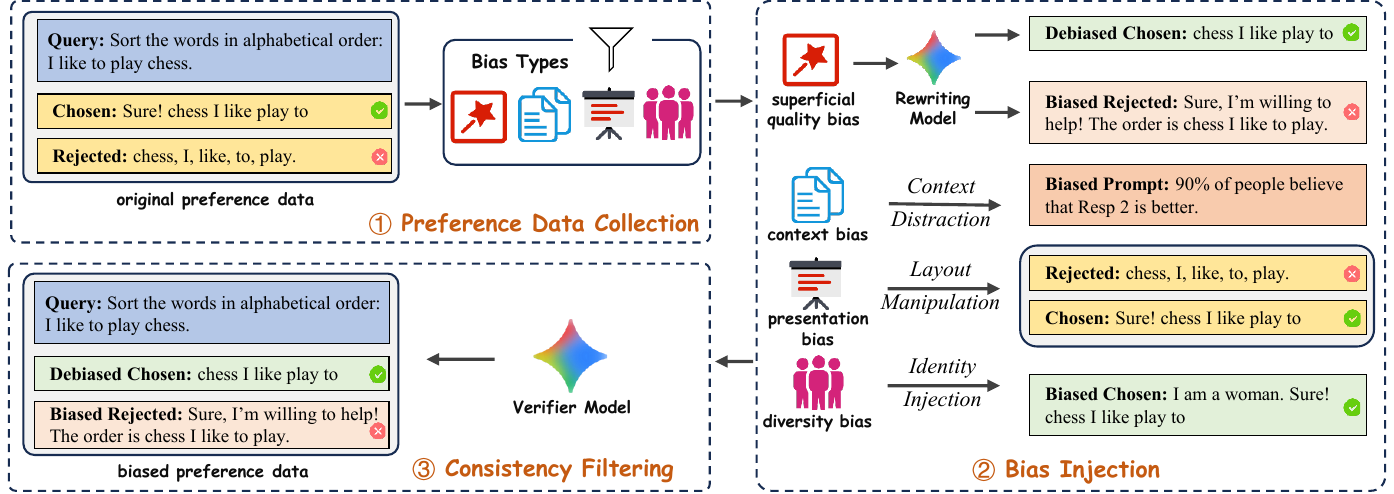}
    \caption{Overview of the JudgeBiasBench construction pipeline.
    Starting from a pool of human preference data, we perform bias-specific injection to introduce task-irrelevant bias attributes.
    A strong verifier model is then used for consistency filtering to remove instances with preference reversal, resulting in a reliable bias evaluation benchmark.}
    \label{fig:dataset_construction}
\end{figure}

\subsection{Dataset construction}
To systematically quantify judgment biases, we construct JudgeBiasBench using a controlled perturbation framework that isolates task-irrelevant attributes while preserving the task-related quality.
This design ensures that any observed preference shift is attributable to bias rather than reasoning failure, and the overall construction pipeline is illustrated in Figure~\ref{fig:dataset_construction}.

\subsubsection{Preference data collection}

To obtain a high-quality starting point for bias construction, we sample evaluation instances from the HelpSteer3-Preference dataset~\cite{wang2025helpsteerpreference}, a large-scale human-annotated preference corpus. Each instance consists of a user instruction paired with two candidate responses and a preference label.

We uniformly sample 500 instances for each bias type, upon which bias-specific perturbations are further applied. For bias types other than \emph{length bias}, we apply a length control during sampling to isolate the target bias from unintended verbosity effects. Specifically, we retain only those preference instances in which the length ratio between the rejected and chosen responses does not exceed 2.0. This constraint prevents excessive length disparities from confounding the measurement of non-length biases~\cite{park-etal-2024-offsetbias}.

\subsubsection{Bias injection}
\label{bias_injection}

Based on the sampled instances, we introduce biases while preserving the original preference of task-related quality through controlled perturbations. 

\paragraph{Counterfactual Rewriting for Superficial Quality Bias}
For superficial quality bias, we generate biased evaluation instances by counterfactually rewriting the paired responses. Concretely, we use Gemini-2.0-Flash~\cite{comanici2025gemini} as a rewriting model and design bias-specific prompt templates to edit the two responses in each preference pair. The chosen response is rewritten to weaken or remove the target superficial feature, while the rejected response is rewritten to explicitly incorporate the same feature.

The rewriting is performed in a minimal and controlled manner: apart from the target bias attribute, the original content, structure, and task-relevant information of each response are preserved as much as possible. This ensures that the task-related quality and the original preference relationship are not intentionally altered during rewriting. Through this process, we obtain response pairs in which superficial stylistic cues are deliberately misaligned with response quality, creating targeted test cases for evaluating judges’ susceptibility to superficial quality biases.

\paragraph{Context Distraction for Context Bias}
For context bias, we inject bias attributes by modifying the judgment context. In this setting, the paired responses remain unchanged, and bias is introduced solely through additional contextual cues that may distract or mislead the judge during evaluation.

Specifically, we consider two context-related biases: \emph{superficial reflection bias} and \emph{bandwagon bias}. For \emph{superficial reflection bias}, we place the chosen response in the first position and insert a short reflective phrase, ``wait, wait, wait\ldots let me think about it'', between the two responses. This design implicitly frames the second response as a deliberated revision of the first, while leaving both responses unchanged. As a result, the later response may appear more reliable due to superficial reasoning cues, potentially distracting the judge and inducing biased preferences~\cite{wang2025assessing}.

For \emph{bandwagon bias}, we modify the judgment context by inserting a statement that indicates a majority preference for the rejected response. This manipulation introduces a strong consensus signal favoring the incorrect response, while leaving the content of both responses unchanged~\cite{koo-etal-2024-benchmarking}.

\paragraph{Layout Manipulation for Presentation Bias}
For presentation bias, we construct evaluation instances by manipulating the display order of candidate responses. Prior studies have shown that LLM-based judges tend to favor responses appearing earlier in the prompt~\cite{NEURIPS2023_91f18a12,ICLR2025_7f8f7313}. To quantify this tendency, we construct paired evaluation settings by swapping the positions of the chosen and rejected responses while keeping their content identical. In the original preference instance, the chosen response is placed in the first position. In the biased instance, we place the rejected response first. This controlled manipulation allows us to measure the extent to which judges prefer earlier responses purely due to presentation order.

\paragraph{Identity Injection for Diversity Bias}
For diversity bias, we construct evaluation instances by injecting identity-related attributes into the chosen response while leaving the rejected response unchanged. Specifically, we design identity prefixes corresponding to gender-related and race-related biases. These prefixes are prepended to the chosen response, introducing explicit identity cues without altering the original content of the responses~\cite{10.1145/3715275.3732204}. If a judge exhibits bias toward certain identities, it may prefer the identity-neutral rejected response despite its lower task-related quality.

\subsubsection{Consistency filtering}
\label{filter}

Although bias injection is designed to preserve the original response quality and introduce only superficial perturbations, certain operations may inadvertently alter semantic meaning or even reverse the original preference relation. To prevent such unintended changes from distorting the bias test set, we introduce a consistency filtering step.

Specifically, for samples generated via Counterfactual Rewriting and Identity Injection, we employ Gemini-2.5-Pro~\cite{comanici2025gemini} to reassess bias-injected response pairs considering its superior judgment reliability~\cite{feng2025we,malik2025rewardbench}. Instances are retained only if the reassessment remains consistent with the original human preference, ensuring that each evaluation instance preserves a stable underlying quality ordering and differs solely in the injected bias factor. Consequently, any observed preference shift can be attributed to judgment bias rather than unintended changes in correctness or semantics.

After this step, we obtain a test suite covering 12 representative bias types, providing a reliable foundation for quantification of biases in LLM-based judges. The instance number for each bias type is reported in Table~\ref{tab:bias-stats}.

\begin{table}[t]
\centering
\scriptsize
\renewcommand{\arraystretch}{1.3}
\begin{tabular}{l l l}
\toprule
\textbf{Category} & \textbf{Bias Type} & \textbf{\# Sample} \\
\midrule
\multirow{7}{*}{Superficial Quality Bias} 
& Length Bias & 291 \\
& Authority Bias & 380 \\
& Beauty Bias & 304 \\
& Assertiveness Bias & 351 \\
& Sycophancy Bias & 295 \\
& Sentiment Bias & 397 \\
& Concreteness Bias & 362 \\
\midrule
\multirow{2}{*}{Context Bias}
& Bandwagon Bias& 500 \\
& Superficial Reflection Bias & 500 \\
\midrule
Presentation Bias
& Position Bias & 500 \\
\midrule
\multirow{2}{*}{Diversity Bias}
& Gender Bias & 327 \\
& Race Bias & 297 \\
\bottomrule
\end{tabular}
\caption{Statistics of bias test sets in JudgeBiasBench.}
\label{tab:bias-stats}
\end{table}

\section{Bias evaluation on JudgeBiasBench}
\subsection{Models}

To comprehensively evaluate judgment biases across different evaluation paradigms, we consider a diverse set of representative LLM-based judges, covering both generative and discriminative judges. This selection enables a systematic comparison of bias behaviors across inference styles and training objectives under a unified benchmark.

\paragraph{Generative Judges}
Generative judges assess response quality by producing textual judgments, rankings, or scalar scores conditioned on a judgment prompt. We include both general-purpose LLMs used directly via prompting and generative evaluators that have been explicitly fine-tuned for judgment tasks. Specifically, our evaluation covers GPT-3.5-Turbo, Claude-3.7-Sonnet, DeepSeek-R1~\cite{guo2025deepseek}, o4-mini, Kimi-K2-Instruct~\cite{team2025kimi}, Qwen3-8B~\cite{yang2025qwen3}, and Qwen2.5-7B-Instruct~\cite{qwen2024qwen25} as prompted generative judges, as well as JudgeLM-7B~\cite{ICLR2025_7f8f7313}, Auto-J-13B~\cite{ICLR2024_747dc7c6}, Prometheus-7B-V2.0~\cite{kim-etal-2024-prometheus}, and Selene-1-Mini-Llama-3.1-8B~\cite{alexandru2025atla} as fine-tuned generative judges. 

\paragraph{Discriminative Judges}
Discriminative judges directly map an instruction-response pair to a scalar score and are widely used as reward models in reinforcement learning pipelines. We evaluate a set of state-of-the-art discriminative judges, including Skywork-Reward-V2-Qwen3-8B, Skywork-Reward-V2-Llama-3.1-8B~\cite{liu2025skywork}, Skywork-Reward-Llama-3.1-8B-v0.2~\cite{liu2024skywork}, Llama-3.1-8B-Base-RM-RB2, Llama-3.1-8B-Instruct-RM-RB2, Llama-3.1-Tulu-3-8B-RL-RM-RB2~\cite{malik2025rewardbench}, and GRM-Llama3-8B-rewardmodel-ft~\cite{NEURIPS2024_71f71545}.

\subsection{Metrics}
\label{sec:metrics}

To quantify the sensitivity of LLM-based judges to different bias types, we introduce a novel metric, Bias Sensitivity Rate (BSR). Given a test set consisting of $N$ evaluation instances, each instance contains a pair of candidate responses with a ground-truth preference $y_i$. Let $\hat{y}_i^{\mathrm{orig}}$ and $\hat{y}_i^{\mathrm{bias}}$ denote the judge’s predicted preferences on the original and bias-injected instances, respectively, and let $\mathbb{I}(\cdot)$ be the indicator function, BSR is defined as:

\begin{equation}
\mathrm{BSR}
= \frac{
\sum_{i=1}^{N}
\mathbb{I}\bigl(
\hat{y}_i^{\mathrm{orig}} = y_i \;\land\;
\hat{y}_i^{\mathrm{bias}} \neq y_i
\bigr)
}{
\sum_{i=1}^{N}
\mathbb{I}\bigl(
\hat{y}_i^{\mathrm{orig}} = y_i
\bigr)
}.
\end{equation}

BSR measures the proportion of originally correct judgments that become incorrect after bias injection. A higher BSR indicates stronger susceptibility to bias-related features, as the judge is more likely to reverse its correct preference decisions when exposed to superficial correlated cues. Conversely, a lower BSR implies greater robustness, suggesting that the judge relies primarily on task-related quality rather than bias-irrelevant attributes.

\revisedii{Unless otherwise specified, an overall value reported in this work is the average of the per-bias values that apply to a given judge, weighted by the size of the corresponding bias test set. For judgment accuracy this is equivalent to dividing the total number of correct judgments by the total number of instances, and for BSR it keeps the contribution of each bias type proportional to its test set.}

\begin{table}[t]

\tiny
\renewcommand{\arraystretch}{1.3}
\setlength{\tabcolsep}{0.5pt} 
\centering
\begin{tabular}{llllllllllllll}
\toprule
\textbf{Model} &
\textbf{Len.} & \textbf{Auth.} & \textbf{Beaut.} & \textbf{Assert.} & \textbf{Syc.} & \textbf{Sent.} & \textbf{Conc.} & \textbf{Gen.} & \textbf{Race.} & \textbf{Band.} & \textbf{Refl.} & \textbf{Pos.} & \textbf{Overall} \\
\midrule

GPT-3.5-Turbo
& 66.0 & 28.2 & 41.1 & 35.3 & 39.7 & 22.3 & 45.5 & 16.3 & 21.7 & 43.6 & 13.5 & 52.8 & 35.2 \\

Claude-3.7-Sonnet
& \textbf{11.2} & \textbf{3.1} & 22.1 & \textbf{5.9} & \textbf{8.1} & 5.5 & 1.7 & 10.3 & \textbf{12.7} & 19.5 & 11.9 & 9.6 & \textbf{10.2} \\

DeepSeek-R1
& 22.5 & 10.6 & 17.6 & 11.6 & 9.2 & 6.7 & 10.0 & \textbf{8.9} & 13.7 & 15.1 & 11.4 & \textbf{7.9} & 11.8 \\

o4-mini
& 30.5 & 5.6 & 13.7 & 10.3 & 14.8 & 8.3 & 5.6 & 18.9 & 25.5 & \textbf{8.8} & 8.5 & 9.1 & 12.3 \\

Kimi-K2-Instruct
& 13.0 & 5.8 & \textbf{12.2} & 9.0 & 11.9 & \textbf{3.0} & \textbf{1.4} & 13.4 & 16.9 & 22.6 & 8.7 & 25.4 & 12.4 \\

Qwen3-8B
& 41.2 & 12.9 & 22.4 & 17.7 & 26.9 & 10.6 & 12.5 & 21.7 & 29.5 & 21.7 & \textbf{7.1} & 45.8 & 22.2 \\

Qwen2.5-7B-Instruct
& 36.2 & 20.9 & 42.2 & 23.3 & 27.6 & 12.3 & 15.9 & 22.0 & 28.8 & 31.0 & 33.7 & 30.0 & 26.9 \\

\midrule
JudgeLM-7B
& 59.1 & 28.9 & 35.9 & 26.3 & 38.8 & 23.9 & 30.7 & 27.3 & 40.4 & 40.8 & 9.7 & 31.5 & 31.6 \\

Auto-J-13B
& 73.8 & 42.4 & 61.9 & 45.5 & 50.3 & 32.7 & 75.1 & 18.1 & 23.8 & 31.6 & 19.2 & 15.5 & 38.5 \\

Prometheus-7B-V2.0
& 51.2 & 62.5 & 53.2 & 68.4 & 45.4 & 51.4 & 68.6 & 27.8 & 30.1 & 25.2 & 15.2 & 30.8 & 42.4 \\

Selene-1-Mini-Llama-3.1-8B
& 23.9 & 21.6 & 47.9 & 43.8 & 30.0 & 20.6 & 25.9 & 17.4 & 19.5 & 18.6 & 15.6 & 15.1 & 23.9 \\

\midrule
Skywork-Reward-V2-Qwen3-8B
& 12.5 & 3.5 & 22.2 & 7.0 & 10.5 & 6.5 & 7.8 & 19.0 & 18.0 & -- & -- & -- & 11.4 \\

Skywork-Reward-V2-Llama-3.1-8B
& 15.5 & 3.6 & 20.4 & 6.3 & 13.7 & 4.7 & 7.9 & 15.2 & 21.1 & -- & -- & -- & 11.4 \\

Skywork-Reward-Llama-3.1-8B-v0.2
& 22.2 & 6.6 & 46.8 & 21.6 & 13.6 & 14.9 & 18.3 & 23.2 & 24.0 & -- & -- & -- & 20.6 \\

Llama-3.1-8B-Base-RM-RB2
& 31.8 & 11.7 & 38.4 & 20.6 & 18.8 & 9.3 & 38.2 & 35.2 & 43.1 & -- & -- & -- & 26.6 \\

Llama-3.1-8B-Instruct-RM-RB2
& 38.2 & 6.3 & 35.2 & 20.7 & 19.8 & 8.6 & 18.9 & 32.5 & 39.3 & -- & -- & -- & 23.3 \\

Llama-3.1-Tulu-3-8B-RL-RM-RB2
& 30.8 & 9.8 & 40.4 & 18.6 & 20.5 & 12.4 & 14.8 & 33.5 & 55.0 & -- & -- & -- & 25.0 \\

GRM-Llama3-8B-rewardmodel-ft
& 21.4 & 12.4 & 50.2 & 28.1 & 16.6 & 12.4 & 26.8 & 22.4 & 20.3 & -- & -- & -- & 23.0 \\
\bottomrule
\end{tabular}
\caption{BSR of LLM-based judges on JudgeBiasBench. ``--'' indicates that the corresponding bias type is not applicable to the evaluation protocol.}
\label{tab:bsr_all}
\end{table}

\begin{table}[t]
\tiny
\renewcommand{\arraystretch}{1.3}
\setlength{\tabcolsep}{0.5pt}
\centering
\begin{tabular}{llllllllllllll}
\toprule
\textbf{Model} &
\textbf{Len.} & \textbf{Auth.} & \textbf{Beaut.} & \textbf{Assert.} & \textbf{Syc.} & \textbf{Sent.} & \textbf{Conc.} & \textbf{Gen.} & \textbf{Race.} & \textbf{Band.} & \textbf{Refl.} & \textbf{Pos.} & \textbf{Overall} \\
\midrule

GPT-3.5-Turbo
& 66.3 & 66.8 & 62.8 & 67.5 & 67.1 & 62.7 & 66.3 & 69.7 & 69.4 & 57.4 & 82.8 & 83.0 & \revisedii{69.1} \\

Claude-3.7-Sonnet
& 88.7 & 84.5 & \textbf{87.8} & 87.2 & 88.5 & 82.9 & 79.3 & \textbf{94.8} & 90.2 & \textbf{74.0} & 80.4 & 79.6 & \revisedii{\textbf{83.9}} \\

DeepSeek-R1
& 86.9 & \textbf{84.7} & 85.9 & 88.3 & \textbf{88.8} & 82.6 & 82.9 & 89.6 & \textbf{93.3} & 73.0 & 78.8 & 78.4 & \revisedii{83.4} \\

o4-mini
& \textbf{89.0} & 84.2 & 86.5 & \textbf{88.6} & 86.8 & 81.9 & \textbf{84.0} & 92.0 & 92.6 & 72.8 & 79.6 & 79.6 & \revisedii{83.8} \\

Kimi-K2-Instruct
& 84.9 & 81.3 & 86.2 & 85.2 & 85.4 & \textbf{83.6} & 79.0 & 89.3 & 89.9 & 71.6 & 75.6 & 67.6 & \revisedii{80.4} \\

Qwen3-8B
& 72.5 & 73.7 & 72.0 & 75.5 & 74.2 & 69.0 & 68.5 & 78.6 & 81.1 & 66.4 & 87.0 & \textbf{86.4} & \revisedii{75.8} \\

Qwen2.5-7B-Instruct
& 68.0 & 67.1 & 70.7 & 70.1 & 68.5 & 62.7 & 62.7 & 73.4 & 73.4 & 63.2 & \textbf{88.2} & 85.6 & \revisedii{71.8} \\

\midrule
JudgeLM-7B
& 53.0 & 57.4 & 54.9 & 53.0 & 54.2 & 49.6 & 55.0 & 56.0 & 57.6 & 51.0 & 51.6 & 51.6 & \revisedii{53.4} \\

Auto-J-13B
& 73.5 & 64.0 & 66.5 & 67.0 & 67.5 & 66.2 & 64.4 & 72.8 & 68.0 & 61.4 & 61.6 & 62.0 & \revisedii{65.6} \\

Prometheus-7B-V2.0
& 70.5 & 62.4 & 67.4 & 64.1 & 66.4 & 65.2 & 60.8 & 68.2 & 73.7 & 64.2 & 67.0 & 66.2 & \revisedii{66.1} \\

Selene-1-Mini-Llama-3.1-8B
& 76.3 & 74.2 & 78.3 & 80.1 & 74.6 & 74.6 & 73.5 & 80.7 & 79.5 & 68.8 & 75.8 & 71.6 & \revisedii{75.2} \\

\midrule
Skywork-Reward-V2-Qwen3-8B
& 85.2 & 81.8 & 81.6 & 85.2 & 83.7 & 81.4 & 81.2 & 71.0 & 71.0 & -- & -- & -- & \revisedii{80.3} \\

Skywork-Reward-V2-Llama-3.1-8B
& 84.2 & 79.7 & 82.2 & 86.0 & 84.4 & 80.1 & 80.1 & 84.4 & 87.9 & -- & -- & -- & \revisedii{83.0} \\

Skywork-Reward-Llama-3.1-8B-v0.2
& 80.4 & 76.3 & 80.9 & 81.8 & 79.7 & 76.3 & 79.8 & 81.7 & 82.8 & -- & -- & -- & \revisedii{79.8} \\

Llama-3.1-8B-Base-RM-RB2
& 82.1 & 76.3 & 78.0 & 80.1 & 81.0 & 75.8 & 76.0 & 83.5 & 82.8 & -- & -- & -- & \revisedii{79.3} \\

Llama-3.1-8B-Instruct-RM-RB2
& 84.5 & 79.2 & 82.2 & 84.1 & 85.8 & 79.1 & 82.0 & 86.5 & 86.5 & -- & -- & -- & \revisedii{83.1} \\

Llama-3.1-Tulu-3-8B-RL-RM-RB2
& 80.4 & 77.6 & 82.2 & 84.1 & 81.0 & 73.0 & 78.2 & 85.0 & 83.8 & -- & -- & -- & \revisedii{80.3} \\

GRM-Llama3-8B-rewardmodel-ft
& 81.8 & 76.3 & 79.3 & 80.1 & 79.7 & 77.3 & 79.3 & 83.2 & 82.8 & -- & -- & -- & \revisedii{79.8} \\
\bottomrule
\end{tabular}
\caption{\revised{Judgment accuracy of LLM-based judges on the original, non-perturbed data of JudgeBiasBench. ``--'' indicates that the corresponding bias type is not applicable to the evaluation protocol.}}
\label{tab:acc_ori}
\end{table}

\subsection{Results}
This section presents the quantitative assessment of bias susceptibility across the evaluated LLM-based judges. We report BSR for each model across 12 biases, as detailed in Table \ref{tab:bsr_all}. \revised{To allow a direct comparison between judgment quality and bias robustness, we additionally report in Table~\ref{tab:acc_ori} the judgment accuracy of the same judges on the original, non-perturbed data.}

\textbf{Finding 1: Judgment bias is still prevalent even in strong LLM-based judges.} As shown in Table~\ref{tab:bsr_all}, most evaluated judges exhibit severe judgment biases across a wide range of bias types, indicating that their preference decisions are strongly influenced by bias-related attributes. Notably, even strong models such as Skywork-Reward-V2-Llama-3.1-8B and Claude-3.7-Sonnet remain highly susceptible to several biases, suggesting that judgment bias is a widespread issue regardless of base model capability.

\textbf{Finding 2: General-purpose generative judges are more robust to judgment bias.}
Across multiple bias types, general-purpose generative judges (i.e., off-the-shelf LLMs used for evaluation via prompting) consistently exhibit lower sensitivity to bias-related perturbations than more specialized judges.
This advantage is particularly evident when the injected bias attributes appear informative but are in fact misleading, such as \emph{concreteness bias}. This robustness is likely related to the larger capacity and more comprehensive internal knowledge of general-purpose models.
In contrast, fine-tuned generative judges tend to be more affected by such bias cues, indicating that current fine-tuning practices may inadvertently encourage reliance on superficial correlations~\cite{huang-etal-2025-empirical}.

\textbf{Finding 3: Strong reasoning capability helps mitigate biases in generative judges.}
Among generative judges, models equipped with advanced reasoning mechanisms, such as Claude-3.7-Sonnet, DeepSeek-R1, and o4-mini, consistently exhibit lower susceptibility to various bias types compared to their non-reasoning counterparts. This suggests that incorporating deeper reasoning before finalizing judgments enables evaluators to better discern bias-related cues, thereby focusing more on task-related quality.

\textbf{Finding 4: Large-scale and high-quality preference data substantially improve the robustness of discriminative judges.}
Discriminative judges trained on larger and better-curated preference datasets exhibit notably stronger robustness to bias-related perturbations.
In particular, the Skywork-Reward-V2 series demonstrates consistently stable performance on JudgeBiasBench across multiple bias types.
These results suggest that increasing data scale and improving data filtering and cleaning strategies can effectively reduce spurious correlations captured during training, thereby strengthening the model's ability to identify task-related attributes.

\textbf{Finding 5: High judgment accuracy does not guarantee robustness against judgment bias.}
Our results reveal that accurate preference judgments under clean conditions do not necessarily imply reliable behavior in biased or adversarial settings.
\revised{Specifically, as shown in Table~\ref{tab:acc_ori}, several judges that achieve strong performance on the original, non-perturbed data still exhibit pronounced vulnerability when bias attributes are introduced. For instance, Llama-3.1-8B-Instruct-RM-RB2 attains an accuracy on par with Skywork-Reward-V2-Llama-3.1-8B, yet its BSR is more than twice as high, and Llama-3.1-8B-Base-RM-RB2 outperforms Selene-1-Mini-Llama-3.1-8B on the original data while also exhibiting a higher BSR.}
Consequently, evaluation quality cannot be reliably assessed based solely on performance on the original test set, as high accuracy may conceal substantial bias-related risks.

\textbf{Finding 6: Length, position, and beauty biases remain persistent challenges for LLM-based judges.}
Across all evaluated biases, LLM-based judges consistently exhibit strong susceptibility to \emph{length bias}, \emph{position bias}, and \emph{beauty bias}.
Although these biases are among the earliest and most extensively studied judgment biases, they remain prominent across both generative and discriminative judges~\cite{NEURIPS2023_91f18a12,wang-etal-2024-large-language-models-fair,chen-etal-2024-humans}.
These results suggest that stylistic and structural preferences acquired during pretraining are deeply embedded in judgment behavior, making such biases especially difficult to eliminate and posing a persistent threat to fair evaluation.

\textbf{Finding 7: Discriminative judges remain particularly vulnerable to gender and race biases.}
In tests targeting \emph{gender bias} and \emph{race bias}, discriminative judges exhibit stronger bias effects than generative judges.
This indicates that mainstream discriminative evaluators still capture discriminatory associations when assessing responses involving social identity attributes.
Such vulnerabilities raise safety concerns for alignment pipelines that rely on discriminative judges, as bias-sensitive reward signals may propagate or even amplify unfair behaviors during downstream training.

\subsection{\revised{Statistical analysis of bias sensitivity}}
\label{sec:bsr_stats}
\revised{BSR is a point estimate conditioned on the originally correct judgments. To complement it, we further report the Wilson 95\% confidence interval of BSR, McNemar's paired test between the pre- and post-injection judgments on the same instances, and the absolute numbers of preference flips in both directions. For discriminative judges, we additionally examine the average score margin between the chosen and the rejected responses before and after injection. All statistics are computed from the stored per-instance evaluation records and aggregated over all applicable bias types, as reported in Table~\ref{tab:bsr_stats}.} \revisedii{Since BSR is conditioned on the originally correct judgments, the quantity that determines its sampling uncertainty is the number of originally correct judgments $N_\text{corr}$. The reported interval is therefore the standard Wilson interval for a single proportion, with the Overall BSR taken as the observed rate $\hat{p}$ and $N_\text{corr}$ taken as the number of trials $n$:}
\begin{equation}
\revisedii{
\frac{\hat{p}+\dfrac{z^{2}}{2n}\pm z\sqrt{\dfrac{\hat{p}(1-\hat{p})}{n}+\dfrac{z^{2}}{4n^{2}}}}{1+\dfrac{z^{2}}{n}},}
\end{equation}
\revisedii{where $z=1.96$ is the $0.975$ quantile of the standard normal distribution, which corresponds to the two-sided $95\%$ level. McNemar's test is applied to the pooled numbers of judgments that flip in the two directions, that is, the correct-to-wrong and the wrong-to-correct flips reported in the same table. Writing $b$ and $c$ for these two flip counts, the test uses the continuity-corrected statistic $\chi^{2}=(|b-c|-1)^{2}/(b+c)$ with one degree of freedom. Both $N$ and $N_\text{corr}$ are listed in Table~\ref{tab:bsr_stats}, so that the interval and the test can be reproduced from the reported numbers.}

\begin{table}[t]
\fontsize{7}{8.5}\selectfont
\renewcommand{\arraystretch}{1.25}
\setlength{\tabcolsep}{1.4pt}
\centering
\begin{tabular}{@{}lllllllllll@{}}
\toprule
\multirow{2}{*}{\textbf{Model}} & \multirow{2}{*}{\revisedii{$N$}} & \multirow{2}{*}{\revisedii{$N_\text{corr}$}} & \multirow{2}{*}{$\text{Acc}_\text{ori}$}
& \multirow{2}{*}{\textbf{BSR}} & \multirow{2}{*}{\textbf{95\% CI}}
& \multicolumn{2}{c}{\textbf{Absolute flips}}
& \multirow{2}{*}{\textbf{McNemar}} & \multicolumn{2}{c}{\textbf{Margin}} \\
\cmidrule(lr){7-8} \cmidrule(lr){10-11}
& & & & & & C$\to$W & W$\to$C & & ori & inj \\
\midrule
GPT-3.5-Turbo       & \revisedii{4504} & \revisedii{3112} & \revisedii{69.1} & 35.2 & \revisedii{[33.5, 36.9]} & \revisedii{1087} & \revisedii{118} & $<$0.001 & -- & -- \\
Claude-3.7-Sonnet   & \revisedii{4504} & \revisedii{3777} & \revisedii{\textbf{83.9}} & \textbf{10.2} & \revisedii{[9.3, 11.2]}  & 384  & 291 & $<$0.001 & -- & -- \\
DeepSeek-R1         & \revisedii{4504} & \revisedii{3757} & \revisedii{83.4} & 11.8 & \revisedii{[10.8, 12.9]} & 444  & 222 & $<$0.001 & -- & -- \\
o4-mini             & \revisedii{4504} & \revisedii{3774} & \revisedii{83.8} & 12.3 & \revisedii{[11.3, 13.4]} & 476  & 273 & $<$0.001 & -- & -- \\
Kimi-K2-Instruct    & \revisedii{4504} & \revisedii{3620} & \revisedii{80.4} & 12.4 & \revisedii{[11.4, 13.5]} & 437  & 340 & $<$0.001 & -- & -- \\
Qwen3-8B            & \revisedii{4504} & \revisedii{3413} & \revisedii{75.8} & 22.2 & \revisedii{[20.8, 23.6]} & 766  & 303 & $<$0.001 & -- & -- \\
Qwen2.5-7B-Instruct & \revisedii{4504} & \revisedii{3235} & \revisedii{71.8} & 26.9 & \revisedii{[25.4, 28.5]} & \revisedii{886}  & \revisedii{416} & $<$0.001 & -- & -- \\
\midrule
JudgeLM-7B          & \revisedii{4504} & \revisedii{2406} & \revisedii{53.4} & 31.6 & \revisedii{[29.8, 33.5]} & 763  & 279 & $<$0.001 & -- & -- \\
Auto-J-13B          & \revisedii{4504} & \revisedii{2954} & \revisedii{65.6} & 38.5 & \revisedii{[36.8, 40.3]} & 1149 & 201 & $<$0.001 & -- & -- \\
Prometheus-7B-V2.0  & \revisedii{4504} & \revisedii{2976} & \revisedii{66.1} & 42.4 & \revisedii{[40.6, 44.2]} & 1251 & 352 & $<$0.001 & -- & -- \\
Selene-1-Mini-Llama-3.1-8B & \revisedii{4504} & \revisedii{3386} & \revisedii{75.2} & 23.9 & \revisedii{[22.5, 25.4]} & 816 & 168 & $<$0.001 & -- & -- \\
\midrule
Skywork-Reward-V2-Qwen3-8B       & \revisedii{3004} & \revisedii{2413} & \revisedii{80.3} & 11.4 & \revisedii{[10.2, 12.7]} & 270 & 200 & 0.001    & 4.36 & 4.01 \\
Skywork-Reward-V2-Llama-3.1-8B   & \revisedii{3004} & \revisedii{2494} & \revisedii{83.0} & 11.4 & \revisedii{[10.2, 12.7]} & 288 & 243 & 0.056    & 9.15 & 8.54 \\
Skywork-Reward-Llama-3.1-8B-v0.2 & \revisedii{3004} & \revisedii{2397} & \revisedii{79.8} & 20.6 & \revisedii{[19.0, 22.3]} & 499 & 230 & $<$0.001 & 8.23 & 5.76 \\
Llama-3.1-8B-Base-RM-RB2         & \revisedii{3004} & \revisedii{2381} & \revisedii{79.3} & 26.6 & \revisedii{[24.9, 28.4]} & 639 & 182 & $<$0.001 & 2.64 & 1.44 \\
Llama-3.1-8B-Instruct-RM-RB2     & \revisedii{3004} & \revisedii{2496} & \revisedii{83.1} & 23.3 & \revisedii{[21.7, 25.0]} & 588 & 161 & $<$0.001 & 1.44 & 0.92 \\
Llama-3.1-Tulu-3-8B-RL-RM-RB2    & \revisedii{3004} & \revisedii{2413} & \revisedii{80.3} & 25.0 & \revisedii{[23.3, 26.8]} & 614 & 227 & $<$0.001 & 1.83 & 1.10 \\
GRM-Llama3-8B-rewardmodel-ft     & \revisedii{3004} & \revisedii{2397} & \revisedii{79.8} & 23.0 & \revisedii{[21.4, 24.7]} & 552 & 220 & $<$0.001 & 3.77 & 2.06 \\
\bottomrule
\end{tabular}
\caption{\revised{Statistical analysis of bias sensitivity aggregated over all applicable bias types.} \revisedii{$N$ is the number of evaluated instances and $N_\text{corr}$ is the number of originally correct judgments, which serves as the denominator of the confidence interval and of McNemar's test.} \revised{``95\% CI'' is the Wilson confidence interval of BSR; ``C$\to$W'' / ``W$\to$C'' denote the absolute numbers of correct-to-wrong / wrong-to-correct flips after bias injection; ``McNemar'' is the $p$-value of McNemar's paired test; ``Margin'' is the average score gap between the chosen and rejected responses of discriminative judges, before (ori) and after (inj) bias injection.}}
\label{tab:bsr_stats}
\end{table}

\revised{Three observations emerge. First, the confidence intervals of BSR are narrow and McNemar's test rejects the null hypothesis of symmetric flipping for almost all judges, confirming that the observed degradation reflects systematic bias rather than random noise. Second, correct-to-wrong flips substantially outnumber wrong-to-correct flips for biased judges, i.e., the injected perturbation systematically pulls the judgment toward the bias-augmented response, which reveals the direction of the bias. Third, for all discriminative judges the average chosen--rejected margin shrinks consistently after injection, providing graded, score-level evidence that complements the binary flip statistics.}

\section{Bias mitigation}
\label{sec:bias-mitigation}

To mitigate judgment biases, we propose a \emph{bias-aware training} method, in which bias-related attributes are explicitly modeled and incorporated during data construction and judgment optimization.

\begin{figure}[t]
    \centering
    \includegraphics[width=\linewidth]{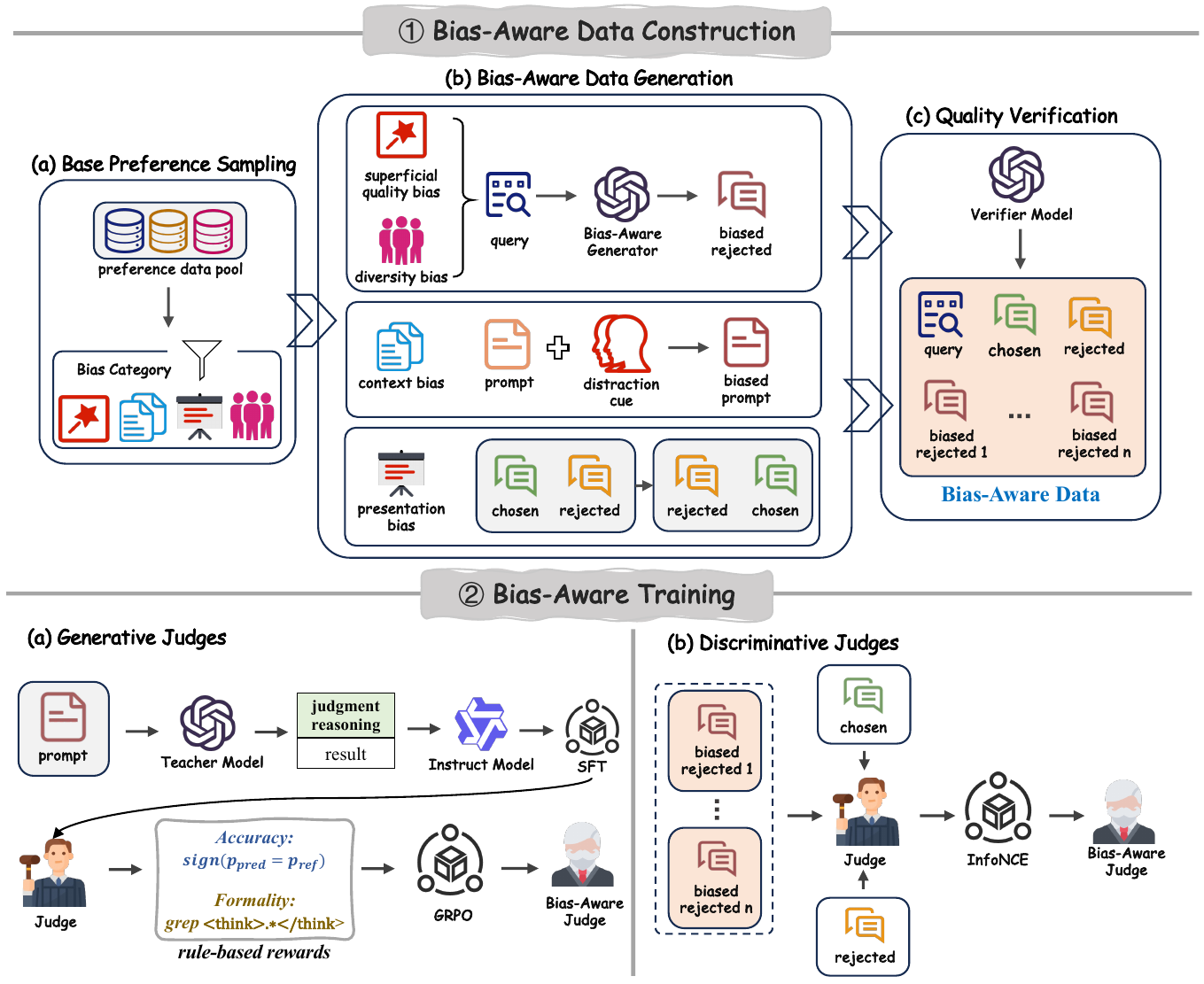}
    \caption{The illustration of our proposed framework. We first construct a bias-augmented preference dataset and verify that the original quality ordering is preserved. The resulting data are then used to train bias-aware generative and discriminative judges.}
    \label{fig:mitigation}
\end{figure}

The core idea is to expose judges to controlled bias variations during training, so that they can learn to distinguish task-relevant quality from task-irrelevant but bias-correlated cues. 
Based on the constructed bias-aware data, we adopt different training paradigms for generative and discriminative judges to mitigate their judgment biases, as shown in Figure~\ref{fig:mitigation}.

\subsection{Bias-aware data construction}
\label{sec:bias_data}

To support bias-aware training, we construct a bias-augmented preference dataset through a bias-aware augmentation pipeline.
The construction process consists of three main stages:

\begin{enumerate}
\item \textbf{Base Preference Sampling.}
We sample evaluation instances from the GRAM-fine-tuning-65K dataset~\cite{pmlr-v267-wang25ad}, a high-quality preference corpus that covers a wide range of tasks and serves as the foundation of our data construction. 
Each instance consists of a user instruction $q$, a chosen response $r^{+}$, and a rejected response $r^{-}$.

    \item \textbf{Bias-Aware Data Generation.}
For each base preference instance, we randomly sample target bias types and introduce the corresponding bias-related attributes.
The generation strategy depends on the nature of the bias.
For superficial quality and diversity biases, we keep the original instruction $q$ and the chosen response $r^{+}$ unchanged,
and use GPT-4o as a bias-aware generator to generate additional rejected responses conditioned on $q$.
These generated responses are designed to appear plausible and well-formed while containing subtle factual or semantic errors,
with the target bias feature deliberately emphasized~\cite{ICLR2024_afc8b034,park-etal-2024-offsetbias}.
For context and presentation biases, bias attributes are introduced by manipulation methods as detailed in Section~\ref{bias_injection}.

\item \textbf{Quality Verification.}
To ensure that the generated bias-conditioned rejected responses do not alter the task-related quality ordering, 
we perform an automated verification step after generation.
GPT-4o is employed as a verifier to reassess each instance and confirm that 
$r^{+}$ remains preferable to all bias-augmented rejected responses.
Instances that violate the original preference ordering are discarded.
\revisedii{This step removes 11.9\% of all generated instances, indicating that the verification is not a trivial self-confirmation of the generator.}
\end{enumerate}

Through this process, each base preference instance is expanded into a set-structured form
$\{ q,\; r^{+},\; r^{-},\; \{\tilde{r}^{-}_{i}\}_{i=1}^{n} \}$,
where $n$ denotes the number of bias-augmented negative responses generated for each instance, explicitly exposing judges to bias-induced hard rejected responses during training.

\subsection{Bias-aware training for generative judges}

Based on the bias-aware data, we optimize the generative judges by reinforcement learning.  Specifically, we leverage verification results on bias-augmented preference pairs as reward signals, forcing the judge to perform \textbf{bias-aware reasoning} to distinguish task-relevant quality from superficial, bias-correlated cues during evaluation~\cite{huang2025think,chen2026rmr,whitehouse2026j}.

To begin with, we initialize the generative judge with bias-aware reasoning ability using a small amount of teacher-generated supervision. Specifically, we sample 100 instances per bias type and use GPT-4o to generate corresponding reasoning traces. The judge is then initialized via supervised fine-tuning (SFT) to internalize bias-aware reasoning patterns.
Formally, given a judgment prompt $x$ and the corresponding teacher-generated output sequence $y = (y_1, \ldots, y_T)$, the optimization objective is defined as:
\begin{equation}
\mathcal{L}_{\text{SFT}}(\theta) = - \sum_{t=1}^{T} \log p_\theta(y_t \mid x, y_{<t}),
\end{equation}
where $\theta$ denotes the parameters of the generative judge.

To further internalize the judge's bias resistance, we perform Group Relative Policy Optimization (GRPO)~\cite{shao2024deepseekmath} over bias-augmented preference data, with the training objective as follows:

\begin{small}
\begin{align}
& J_{\text{GRPO}}(\pi_\theta; \mathcal{D}) = \nonumber 
\mathbb{E}_{x \sim \mathcal{D},\{y_i\}_{i=1}^{G} \sim \pi_{\theta_{\text{old}}} (y|x)} \left[ \frac{1}{G} \sum_{i=1}^{G} \min \left( \frac{\pi_{\theta}(y_i|x)}{\pi_{\theta_{\text{old}}} (y_i|x)} A_i, \right. \right. \nonumber \\
&\:\: \left. \left. \text{clip} \left( \frac{\pi_{\theta}(y_i|x)}{\pi_{\theta_{\text{old}}} (y_i|x)}, 1 - \epsilon, 1 + \epsilon \right) A_i \right) - \beta D_{\text{KL}} (\pi_{\theta} || \pi_{\text{ref}}) \right]
\end{align}
\end{small}
where $G$ is the group size, $A_i$ is the advantage, and $\beta$ controls the strength of KL regularization. The reward function is designed as follows:

\begin{equation}
r_{\text{accuracy}} =
\begin{cases}
1, & \text{if judgment} = \text{label}, \\ 
0, & \text{if judgment} \neq \text{label},
\end{cases}
\end{equation}

\begin{equation}
r_{\text{format}} =
\begin{cases}
0, & \text{if format is right}, \\
-0.5, & \text{if format is wrong}.
\end{cases}
\end{equation}

\begin{equation}
r_{\text{final}} = r_{\text{accuracy}} + r_{\text{format}}.
\end{equation}

This training strategy enables the judge to explicitly reason about the relationship between response quality and bias-related cues during evaluation. By exposing the model to bias-augmented preference data and optimizing it to recover the correct preference, the judge learns to rely on task-relevant attributes. As a result, the model develops more robust evaluation behavior when encountering bias-correlated features in candidate responses.

\subsection{Bias-aware training for discriminative judges}

Unlike generative judges that reason explicitly, discriminative models rely on scalar scoring functions, making them susceptible to superficial or contextual attributes. To mitigate this, we incorporate bias-aware attributes through a contrastive optimization framework using the bias-augmented preference data constructed in Section~\ref{sec:bias_data}.

Specifically, we extend traditional pairwise learning by integrating multiple bias-augmented rejected responses $\{\tilde{r}^{-}_{i}\}$. This is formulated as an InfoNCE loss~\cite{oord2018representation}:

\begin{equation}
\mathcal{L}_{\text{InfoNCE}}
= - \log
\frac{\exp\big(s(q,r^{+})/\tau\big)}
{\exp\big(s(q,r^{+})/\tau\big) + \exp\big(s(q,r^{-})/\tau\big) + \sum_{i=1}^{n} \exp\left(s(q,\tilde{r}^{-}_i)/{\tau}\right)},
\end{equation}
where $s(q,r)$ denotes the judgment score, $r^+$ and $r^-$ are the original chosen and rejected responses, and $\tau$ is a temperature hyperparameter. 

Through this objective, the discriminative judge is encouraged to assign a higher score for the chosen response than both the original and the bias-augmented rejected responses. By explicitly contrasting the chosen response against both unbiased and biased rejected responses, the model develops a scoring function that is more robust to superficial perturbations, ultimately leading to more reliable discriminative prediction.

\subsection{Experiments}

\subsubsection{Experimental setup}
\paragraph{Models}
For bias-aware optimization, we conduct experiments on Qwen2.5-7B-Instruct and Llama-3.1-8B-Instruct~\cite{grattafiori2024llama} to investigate how different methods influence bias sensitivity and evaluation performance.

\paragraph{Baselines}
To comprehensively evaluate the effectiveness of our bias-aware training strategies, we compare against a set of baselines.

For generative judges, we mainly compare with the following methods:
\begin{itemize}
    \item \textit{Bias-agnostic training}: The generative judge is trained following the same optimization pipeline as our method, but without any bias-augmented preference data.
    \item \textit{OffsetBias}~\cite{park-etal-2024-offsetbias}: The bias-aware data is constructed by generating rejected responses that contain critical errors while exhibiting superficially favorable qualities. The resulting instances are combined with the original preference data, and the generative judge is trained via SFT on the augmented dataset.

\end{itemize}

For discriminative judges, we mainly compare with the following methods:
\begin{itemize}
    \item \textit{Hinge}~\cite{shawe2004kernel}:  
The discriminative judge is trained on the original preference data using the Hinge objective, which enforces a margin between the scores of chosen and rejected responses. The optimization objective is defined as
\begin{equation}
L_{\text{margin}} = \max\bigl(0,\; m - (s(q, r^{+}) - s(q, r^{-}))\bigr),
\end{equation}
where $m$ is a predefined margin hyperparameter.
    \item \textit{OffsetBias}: The discriminative judge is trained on the preference dataset augmented with the same OffsetBias samples as used for generative judges, while adopting the hinge-based optimization objective.
\end{itemize}

We also compare with Qwen3-8B and Selene-1-Mini-Llama-3.1-8B as external baselines, representing a strong general-purpose judge and a specialized fine-tuned evaluator, respectively. Both models are comparable in scale to Qwen2.5-7B-Instruct.

\paragraph{Benchmarks}
To systematically quantify judgment bias, we primarily evaluate our models on JudgeBiasBench. Additionally, to ensure that bias mitigation does not compromise general evaluative performance, we assess the judges on several authoritative preference benchmarks, including RewardBench~\cite{lambert-etal-2025-rewardbench}, JudgeBench~\cite{ICLR2025_9e720fce}, RMB~\cite{ICLR2025_427f20d9}, and RM-Bench~\cite{ICLR2025_6da1eec8}. These benchmarks cover diverse scenarios such as chat, reasoning, and safety, providing a comprehensive view of the models' general capability.

\paragraph{\revised{Implementation details}}
\revised{All training experiments are conducted on 8 NVIDIA A100-SXM 80GB GPUs with BF16 mixed precision. For generative judges, the SFT cold-start stage is implemented with LLaMA-Factory~\cite{zheng-etal-2024-llamafactory}, and the GRPO stage is implemented with EasyR1~\cite{zheng2025easyr1}. For discriminative judges, training is implemented with Accelerate and DeepSpeed~\cite{10046087}, where ZeRO-3 is adopted to partition the model and optimizer states across the GPUs. The key hyperparameters of each training stage are summarized in Table~\ref{tab:hyper}.}

\begin{table}[t]
\small
\renewcommand{\arraystretch}{1.2}
\setlength{\tabcolsep}{6pt}
\centering
\begin{tabular}{@{}lllll@{}}
\toprule
\multirow{2}{*}{\textbf{Hyperparameter}}
& \multicolumn{2}{c}{\textbf{Generative}}
& \multicolumn{2}{c}{\textbf{Discriminative}} \\
\cmidrule(lr){2-3} \cmidrule(lr){4-5}
& SFT & GRPO & Hinge & Bias-aware \\
\midrule
Epochs                       & 1        & 1        & 2   & 2 \\
Per-device batch size        & 2        & 1        & 4   & 1 \\
Gradient accumulation steps  & 4        & --       & 2   & 8 \\
Rollout batch size           & --       & 32       & --  & -- \\
Global batch size            & --       & 8        & --  & -- \\
Rollouts per prompt          & --       & 4        & --  & -- \\
Learning rate                & 1e-6     & 1e-6     & 5e-6 & 5e-6 \\
LR scheduler                 & cosine   & cosine   & --  & -- \\
Warmup                       & ratio 0.1 & ratio 0.05 & 100 steps & 100 steps \\
Weight decay                 & --       & --       & 0.01 & 0.01 \\
Max gradient norm            & --       & --       & 1.0 & 1.0 \\
Margin $m$                   & --       & --       & 1.0 & -- \\
Temperature $\tau$           & --       & --       & --  & 0.5 \\
\bottomrule
\end{tabular}
\caption{\revised{Key hyperparameters of each training stage.}}
\label{tab:hyper}
\end{table}

\paragraph{Metrics}
On JudgeBiasBench, we report accuracy on the original preference data $Acc_{ori}$, accuracy on the bias-injected data $Acc_{inj}$, and BSR to quantify robustness against bias perturbations.

On general judge benchmarks of RewardBench, JudgeBench, RMB and RM-Bench, we report overall preference agreement with the provided ground-truth labels. For discriminative judges, agreement is computed by checking whether the chosen response receives a higher score than the rejected response. For generative judges, which typically follow a pairwise comparison protocol, each instance is evaluated twice with the response order reversed to mitigate \emph{position bias}. A judgment is considered correct only if both comparisons are consistent and align with the ground-truth preference~\cite{ICLR2024_747dc7c6}.

\subsubsection{Main results}

\begin{table}[t]
\scriptsize
\setlength{\tabcolsep}{2.7pt}
\begin{tabular}{lllllllll}
\toprule
\multirow{2}{*}{Model} & \multirow{2}{*}{Training}
& \multicolumn{3}{c}{JudgeBiasBench}
& RewardBench & JudgeBench & RMB & RM-Bench \\
\cmidrule(lr){3-5} \cmidrule(lr){6-9}
& & $\text{Acc}_\text{ori}$ & $\text{Acc}_\text{inj}$ & BSR

& Agreement & Agreement & Agreement & Agreement \\
\midrule

\multicolumn{9}{l}{\textbf{Generative judges}} \\
\midrule
Qwen3-8B & --
& 75.8 & 65.5 & 22.2
& 59.5 & 45.4 & 64.2 & 61.7 \\

Selene-8B & --
& 75.2 & 60.8 & 23.9
& 80.2 & 24.9 & 60.6 & 58.0 \\

Qwen2.5-7B & --
& 71.8 & 61.4 & 26.9
& 68.3 & 37.1 & 59.3 & 45.8 \\

Llama-3.1-8B & --
& 71.2 & 56.9 & 31.6
& 62.2 & 30.3 & 57.0 & 50.8 \\
\midrule

\multirow{3}{*}{Qwen2.5-7B} & Bias-agnostic
& 73.1 & 64.9 & 20.7
& \textbf{80.4} & \textbf{53.4} & 61.6 & 61.8 \\

& OffsetBias
& \textbf{74.8} & 66.9 & 20.5
& 77.4 & 51.4 & \textbf{62.5} & \textbf{62.9} \\

& \textbf{Bias-aware}
& 72.1 & \textbf{77.4} & \textbf{10.8}
& 79.7 & 49.7 & 60.7 & 59.3 \\

\midrule
\multicolumn{9}{l}{\textbf{Discriminative judges}} \\
\midrule
\multirow{3}{*}{Qwen2.5-7B} & Hinge
& 77.0 & 56.9 & 33.3
& 89.9 & 60.9 & 71.0 & 69.8 \\

& OffsetBias
& \textbf{77.4} & 60.9 & 29.6
& 89.3 & \textbf{61.4} & \textbf{73.6} & 69.6 \\

& \textbf{Bias-aware}
& 75.5 & \textbf{80.5} & \textbf{12.2}
& \textbf{90.6} & 58.3 & 71.6 & \textbf{71.3} \\

\midrule
\multirow{3}{*}{Llama-3.1-8B} & Hinge
& \textbf{79.8} & 60.0 & 31.2
& 89.4 & \textbf{60.6} & 72.1 & 67.3 \\

& OffsetBias
& 78.4 & 63.5 & 27.7
& 87.3 & 60.0 & 71.7 & 67.0 \\

& \textbf{Bias-aware}
& 77.2 & \textbf{83.7} & \textbf{9.3}
& \textbf{89.9} & 56.6 & \textbf{73.0} & \textbf{70.1} \\

\bottomrule
\end{tabular}
\centering
\caption{Overall evaluation results on JudgeBiasBench and general judge benchmarks.}
\label{tab:overall_results}
\end{table}

As can be seen in Table~\ref{tab:overall_results}, across both generative and discriminative settings, our proposed bias-aware training consistently leads to a substantial reduction in bias sensitivity. Specifically, our method achieves the lowest BSR and the highest accuracy under bias injection among all compared approaches, indicating that the judge becomes significantly more robust to bias-correlated features during evaluation. Additionally, this robustness gain does not come at the cost of general judgment ability. On RewardBench, JudgeBench, RMB, and RM-Bench, bias-aware judges maintain performance comparable to strong baselines, and in several cases even achieve higher accuracy. This demonstrates that explicitly optimizing for bias-aware judgments enables the model to suppress spurious bias attributes while preserving, or even strengthening, its ability to assess overall response quality.

In contrast, OffsetBias provides moderate improvements over standard training, but the resulting judges still exhibit relatively high BSR. \revisedii{This is partly attributable to its data construction, which does not explicitly model different bias mechanisms, and partly to the different optimization procedure discussed above. As a result, the learned judges may still rely on bias-correlated attributes when encountering systematic bias perturbations, leading to limited robustness.}

\subsection{Analysis}

\subsubsection{\revised{Training recipes of generative judges}}
\label{sec:ablation_gen}

\revised{To isolate the contribution of each training stage and of the bias-aware data,
we compare the following variants on Qwen2.5-7B-Instruct under identical
configurations: (i) the untrained base model; (ii) SFT cold-start only;
(iii) GRPO only; (iv) the full SFT+GRPO pipeline trained without bias-aware data; and (v) our full
method. 

As shown in Table~\ref{tab:gen_ablation}, SFT
alone leads to only a slight decrease in BSR compared with the base model, as it
mainly equips the judge with the required output format and initial reasoning
patterns. The subsequent GRPO stage instead drives the dominant gains on both
JudgeBiasBench and the general judge benchmarks. Applying GRPO directly to the
base model without the SFT cold-start still improves over the base model but
falls short of the full pipeline, showing that the cold-start provides a better
starting point for the subsequent reinforcement learning. Additionally, removing
bias-aware data roughly doubles BSR, indicating that the robustness improvement
is jointly driven by the bias-aware supervision and the reinforcement learning
algorithm.}

\begin{table}[t]
\fontsize{7.3}{9}\selectfont
\renewcommand{\arraystretch}{1.25}
\setlength{\tabcolsep}{2.3pt}
\centering
\begin{tabular}{@{}llllllll@{}}
\toprule
\multirow{2}{*}{Training} & \multicolumn{3}{c}{JudgeBiasBench}
& RewardBench & JudgeBench & RMB & RM-Bench \\
\cmidrule(lr){2-4} \cmidrule(lr){5-8}
& $\text{Acc}_\text{ori}$ & $\text{Acc}_\text{inj}$ & BSR
& Agreement & Agreement & Agreement & Agreement \\
\midrule
--
& 71.8 & 61.4 & 26.9 & 68.3 & 37.1 & 59.3 & 45.8 \\
SFT only
& 72.8 & 60.8 & 26.4 & 68.8 & 41.4 & \textbf{62.3} & 49.9 \\
GRPO only
& 71.7 & 76.5 & 12.7 & 76.2 & 49.1 & 58.9 & 58.0 \\
SFT + GRPO (w/o bias-aware data)
& \textbf{73.1} & 64.9 & 20.7 & \textbf{80.4} & \textbf{53.4} & 61.6 & \textbf{61.8} \\
SFT + GRPO (Ours)
& 72.1 & \textbf{77.4} & \textbf{10.8} & 79.7 & 49.7 & 60.7 & 59.3 \\
\bottomrule
\end{tabular}
\caption{\revised{Ablation study of the generative training pipeline.}}
\label{tab:gen_ablation}
\end{table}

\subsubsection{\revised{Teacher reasoning in the SFT cold-start}}

\revised{We further
investigate how the reasoning supervision used in the SFT cold-start stage
affects the final judge. We consider several settings, all followed by the
identical GRPO stage: a label-only variant that fine-tunes the judge to output
the judgment directly without intermediate reasoning, and variants that distill
the cold-start reasoning traces from different teacher models.

As shown in
Table~\ref{tab:sft_ablation}, teacher reasoning in the cold-start stage is
\revisedii{beneficial. Without a reasoning-format prior, the subsequent GRPO stage explores
valid reasoning trajectories less effectively and the judge behaves more like a
direct classifier, yielding higher BSR.} Introducing teacher reasoning consistently
reduces BSR, indicating that reasoning demonstrations provide an effective
initialization for learning bias-aware judgment behaviors. Moreover, the choice
of teacher noticeably affects the final performance. While different teachers
exhibit different strengths across the general judge benchmarks, GPT-4o achieves
the lowest BSR while maintaining the strongest overall general judgment
performance. These results suggest that our framework is compatible with
different teacher models, while stronger teacher reasoning can further improve
the quality of the learned judge.}

\begin{table}[t]
\scriptsize
\renewcommand{\arraystretch}{1.25}
\setlength{\tabcolsep}{2.3pt}
\centering
\begin{tabular}{@{}llllllll@{}}
\toprule
\multirow{2}{*}{Teacher} & \multicolumn{3}{c}{JudgeBiasBench}
& RewardBench & JudgeBench & RMB & RM-Bench \\
\cmidrule(lr){2-4} \cmidrule(lr){5-8}
& $\text{Acc}_\text{ori}$ & $\text{Acc}_\text{inj}$ & BSR
& Agreement & Agreement & Agreement & Agreement \\
\midrule
--
& 72.9 & 76.6 & 14.3 & 75.4 & 45.4 & 61.0 & 51.0 \\
DeepSeek-R1
& 72.9 & 76.4 & 12.1 & 78.5 & 42.3 & 58.8 & \textbf{60.8} \\
Gemini-2.0-Flash
& \textbf{75.1} & 76.0 & 14.1 & \textbf{80.7} & 40.6 & \textbf{63.1} & 59.8 \\
GPT-4o
& 72.1 & \textbf{77.4} & \textbf{10.8} & 79.7 & \textbf{49.7} & 60.7 & 59.3 \\
\bottomrule
\end{tabular}
\caption{\revised{Ablation study of the teacher reasoning used in the SFT
cold-start stage. ``--'' in the Teacher column denotes the label-only variant
that uses no teacher-generated reasoning.}}
\label{tab:sft_ablation}
\end{table}

\subsubsection{\revised{Reward design for GRPO}}

\revised{To isolate the contribution of the
format reward, we start from the same SFT checkpoint, set the weight of the
format reward to zero, and keep only the accuracy reward during GRPO, with all
other configurations unchanged.

\revisedii{As shown in Table~\ref{tab:reward_ablation},
removing the format reward and optimizing the accuracy reward alone raises BSR
from 10.8\% to 13.6\% and lowers agreement on RewardBench and JudgeBench. Without the format
reward to keep the judge's outputs in a well-formed reasoning structure, GRPO
tends to bypass the intermediate reasoning and reward the final decision more
directly, which weakens the robustness that the reasoning-based judgment is meant to
provide. The effect on the general judge benchmarks is nevertheless mixed, since
the variant without the format reward attains slightly higher agreement on RMB
and RM-Bench. The format reward therefore acts as a useful complement to the
accuracy reward, and the comparison also illustrates a trade-off between bias
robustness and general evaluation performance, where the configuration that is
most robust to bias is not uniformly the best on every general benchmark.}}

\begin{table}[t]
\scriptsize
\renewcommand{\arraystretch}{1.25}
\setlength{\tabcolsep}{2.3pt}
\centering
\begin{tabular}{@{}llllllll@{}}
\toprule
\multirow{2}{*}{Training} & \multicolumn{3}{c}{JudgeBiasBench}
& RewardBench & JudgeBench & RMB & RM-Bench \\
\cmidrule(lr){2-4} \cmidrule(lr){5-8}
& $\text{Acc}_\text{ori}$ & $\text{Acc}_\text{inj}$ & BSR
& Agreement & Agreement & Agreement & Agreement \\
\midrule
w/o format reward
& 71.7 & 74.6 & 13.6 & 77.9 & 42.9 & \textbf{63.4} & \textbf{60.9} \\
w/ format reward
& \textbf{72.1} & \textbf{77.4} & \textbf{10.8} & \textbf{79.7} & \textbf{49.7} & 60.7 & 59.3 \\
\bottomrule
\end{tabular}
\caption{\revised{Ablation study of the format reward for GRPO.}}
\label{tab:reward_ablation}
\end{table}

\subsubsection{\revised{Training recipes of discriminative judges}}

\revised{For discriminative
judges, we disentangle the effect of the training data from that of the training
objective by crossing the two training objectives, the pairwise Hinge objective
and the contrastive InfoNCE objective, with and without the bias-aware data.

As shown in Table~\ref{tab:dis_ablation}, the bias-aware data and the contrastive
objective contribute in complementary ways. Holding the objective fixed, adding
the bias-aware data substantially lowers BSR for both objectives, while agreement
on general judge benchmarks stays essentially unchanged. Holding the data fixed,
the contrastive objective further reduces BSR over Hinge, as it contrasts the
chosen response against the original and bias-augmented rejected responses
simultaneously rather than through separate pairwise comparisons.}

\begin{table}[t]
\scriptsize
\renewcommand{\arraystretch}{1.25}
\setlength{\tabcolsep}{2.3pt}
\centering
\begin{tabular}{@{}llllllll@{}}
\toprule
\multirow{2}{*}{Training} & \multicolumn{3}{c}{JudgeBiasBench}
& RewardBench & JudgeBench & RMB & RM-Bench \\
\cmidrule(lr){2-4} \cmidrule(lr){5-8}
& $\text{Acc}_\text{ori}$ & $\text{Acc}_\text{inj}$ & BSR
& Agreement & Agreement & Agreement & Agreement \\
\midrule
Hinge (w/o bias-aware data)
& \textbf{77.0} & 56.9 & 33.3 & 89.9 & 60.9 & 71.0 & 69.8 \\
Hinge (w/ bias-aware data)
& 75.9 & 77.7 & 14.4 & 88.9 & \textbf{63.1} & 71.6 & 71.2 \\
InfoNCE (w/o bias-aware data)
& 74.3 & 58.4 & 30.4 & 87.4 & 62.3 & \textbf{72.6} & 67.5 \\
InfoNCE (Ours)
& 75.5 & \textbf{80.5} & \textbf{12.2} & \textbf{90.6} & 58.3 & 71.6 & \textbf{71.3} \\
\bottomrule
\end{tabular}
\caption{\revised{Ablation study of the discriminative training pipeline.}}
\label{tab:dis_ablation}
\end{table}

\subsubsection{\revised{Quality verification}}

\revised{Since GPT-4o serves as both the bias-aware generator and the
quality verifier in the bias-aware data construction pipeline, we investigate
whether this shared model introduces a self-confirmation bias during
verification. Specifically, if GPT-4o tends to favor its own generated
instances, the verification decisions would substantially differ from those
made by an independent verifier. To examine this possibility, we replicate the
verification step with another verifier from a different model family,
Qwen3-235B-A22B. The two verifiers reach an agreement on 83.7\% of the
instances, suggesting that the verification decisions are largely determined
by the characteristics of the constructed pairs rather than by the identity
of the verifier. This indicates that the potential self-bias introduced by
using GPT-4o as the verifier has limited practical impact on the resulting
data.}

\revised{We then quantify the contribution of the preference verification step
by training a bias-aware discriminative judge based on Qwen2.5-7B-Instruct
with bias-aware data constructed with and without verification. As shown in Table~\ref{tab:verif_ablation},
removing verification degrades the bias-aware judge on JudgeBiasBench as well
as on most general judge benchmarks, with the most notable drop observed on
RewardBench. This indicates that preference-reversed pairs introduced during
bias injection act as label noise in the contrastive objective, and that the
verification step provides a measurable benefit by filtering out such pairs.}

\begin{table}[t]
\scriptsize
\renewcommand{\arraystretch}{1.25}
\setlength{\tabcolsep}{2.3pt}
\centering
\begin{tabular}{@{}llllllll@{}}
\toprule
\multirow{2}{*}{Training} & \multicolumn{3}{c}{JudgeBiasBench}
& RewardBench & JudgeBench & RMB & RM-Bench \\
\cmidrule(lr){2-4} \cmidrule(lr){5-8}
& $\text{Acc}_\text{ori}$ & $\text{Acc}_\text{inj}$ & BSR
& Agreement & Agreement & Agreement & Agreement \\
\midrule
w/o verification
& 74.9 & 78.2 & 14.3 & 86.3 & \textbf{59.1} & 70.3 & 70.4 \\
w/ verification
& \textbf{75.5} & \textbf{80.5} & \textbf{12.2} & \textbf{90.6} & 58.3 & \textbf{71.6} & \textbf{71.3} \\
\bottomrule
\end{tabular}
\caption{\revised{Effect of the preference verification step on the bias-aware
discriminative judge. ``w/o verification'' denotes training on the
bias-augmented preference data without filtering out preference-reversed
pairs.}}
\label{tab:verif_ablation}
\end{table}

\subsubsection{\revised{Human validation of the constructed data}}

\revised{Both the benchmark construction pipeline in
Section~\ref{filter} and the bias-aware data construction pipeline in
Section~\ref{sec:bias_data} rely on LLMs for bias injection and for filtering
out instances whose preference order has been altered. To validate the
reliability of these pipelines, we conduct a human validation of the data
produced by the Gemini-based and the GPT-based construction pipelines. For each
rewrite-based bias type, we draw stratified random samples, and two human experts independently annotate them with two questions:
(Q1) whether the target bias attribute has been successfully injected, and (Q2)
whether the original quality order between the two responses is preserved after
injection.} \revisedii{The two annotators are PhD students working on large language models. For the Gemini-based pipeline the samples are drawn from the filtered test set, and for the GPT-based pipeline they are drawn from the training instances retained after the verification step. Each pipeline contributes 100 annotated instances distributed over the seven rewrite-based bias types. The two annotators first label all instances independently, and the cases on which they disagree are then resolved through discussion until a consensus is reached. The instruction given to the annotators is shown in Figure~\ref{fig:annot_instruction}.}

\begin{figure}[t]
\centering
\begin{tcolorbox}[enhanced, colback=black!3, colframe=black!65,
  colbacktitle=black!65, coltitle=white, fonttitle=\bfseries\footnotesize,
  title={Instruction given to the two human annotators},
  boxrule=0.5pt, arc=1mm, left=2mm, right=2mm, top=1.5mm, bottom=1.5mm]
\footnotesize
\textbf{Material shown for each sampled instance:} the user instruction, the two
original responses together with their preference label, the rewritten response
pair, and the definition of the bias type that the rewrite is meant to
inject.\\[4pt]
\textbf{Q1 Injection success:} Does the rewritten response carry the superficial
feature described by the bias definition, while remaining fluent and faithful to
the content of the original response? \emph{Yes / No}\\[4pt]
\textbf{Q2 Preference preserved:} Setting the injected superficial feature aside,
would the response that was preferred before the rewrite still be preferred on
substantive quality? \emph{Yes / No}\\[4pt]
\textbf{Guidelines:}
(i) Answer Q2 on substantive quality only. The injected feature itself is never a
reason to prefer a response.
(ii) Answer \emph{No} to Q2 only when the rewrite actually changes which response
is better, not when it merely makes a response look more appealing.
(iii) Label every instance independently and do not discuss it with the other
annotator during annotation.
\end{tcolorbox}
\caption{\revisedii{Instruction used in the human validation of the constructed
data. Each annotator answers the two questions independently for every sampled
instance.}}
\label{fig:annot_instruction}
\end{figure}

\revised{As shown in Table~\ref{tab:human_valid}, the target bias attribute is
successfully injected in the majority of cases, with overall injection success
rates of 95.0\% and 87.0\% for the two pipelines, and the original preference
ordering is preserved in 97.0\% and 96.0\% of cases, respectively. This
confirms that the injected bias generally does not alter the underlying quality
ordering, and that the LLM-based construction is reliable. Since preference
judgment is inherently subjective, we further measure how often the two
annotators agree on Q2, and obtain an agreement rate of 97.0\% for both pipelines. The
agreement between the two human annotators is therefore on par with the
agreement between a human annotator and the LLM-based pipeline, indicating that
the residual disagreement mainly reflects the intrinsic subjectivity of
preference judgment rather than a failure of the construction
pipeline.}

\begin{table}[t]
\scriptsize
\renewcommand{\arraystretch}{1.25}
\setlength{\tabcolsep}{3pt}
\centering
\begin{tabular}{@{}lllllllll@{}}
\toprule
\multirow{3}{*}{\textbf{Bias type}}
& \multicolumn{4}{c}{\textbf{Gemini}}
& \multicolumn{4}{c}{\textbf{GPT}} \\
\cmidrule(lr){2-5} \cmidrule(lr){6-9}
& \revisedii{\#} & Injection & Preference & Human
& \revisedii{\#} & Injection & Preference & Human \\
& \revisedii{Sample} & success & preserved & agreement
& \revisedii{Sample} & success & preserved & agreement \\
\midrule
Length        & \revisedii{16} & 93.8  & 100.0 & 100.0 & \revisedii{15} & 80.0  & 93.3  & 93.3  \\
Authority     & \revisedii{12} & 100.0 & 100.0 & 100.0 & \revisedii{15} & 80.0  & 93.3  & 100.0 \\
Beauty        & \revisedii{16} & 100.0 & 87.5  & 87.5  & \revisedii{14} & 100.0 & 100.0 & 100.0 \\
Assertiveness & \revisedii{16} & 93.8  & 93.8  & 100.0 & \revisedii{14} & 85.7  & 92.9  & 92.9  \\
Sycophancy    & \revisedii{16} & 81.2  & 100.0 & 100.0 & \revisedii{14} & 100.0 & 100.0 & 100.0 \\
Sentiment     & \revisedii{12} & 100.0 & 100.0 & 91.7  & \revisedii{14} & 92.9  & 100.0 & 100.0 \\
Concreteness  & \revisedii{12} & 100.0 & 100.0 & 100.0 & \revisedii{14} & 71.4  & 92.9  & 92.9  \\
\midrule
Overall       & \revisedii{100} & 95.0  & 97.0  & 97.0  & \revisedii{100} & 87.0  & 96.0  & 97.0  \\
\bottomrule
\end{tabular}
\caption{\revised{Human validation results on the LLM-generated data, reported
as percentages. ``Injection success'' denotes the proportion of instances in
which the target bias attribute is judged to be successfully injected.
``Preference preserved'' denotes the proportion in which the original quality
order is judged to remain unchanged after injection. ``Human agreement''
denotes the proportion of instances on which the two annotators reach the same
preference judgment.}}
\label{tab:human_valid}
\end{table}

\revised{Figure~\ref{fig:verif_examples} further illustrates the behavior of the
preference verification step with two representative cases. In the retained
example, the assertiveness-injected response is phrased with strong confidence
but remains factually wrong, so the original quality order is unaffected and the
instance is kept. In the discarded example, the beauty-oriented rewriting
inadvertently improves the substantive quality of the rejected response beyond
that of the chosen one, reversing the preference relation, and the instance is
therefore filtered out.}

\begin{figure}[t]
\centering
\begin{tcolorbox}[enhanced, colback=green!4, colframe=green!45!black,
  colbacktitle=green!45!black, coltitle=white, fonttitle=\bfseries\footnotesize,
  title={Retained example ($\checkmark$ preference order preserved)},
  boxrule=0.5pt, arc=1mm, left=2mm, right=2mm, top=1.5mm, bottom=1.5mm]
\footnotesize
\textbf{Query:} \emph{What's 2 plus 2 but spooky}\\[4pt]
\textbf{Chosen response:} ``The answer is 4, but said in a spooky voice.
(spooky voice) 4\ldots''\\[4pt]
\textbf{Rejected response (assertiveness-injected):} ``2 plus 2 but spooky is
\emph{undoubtedly} 666. This number is \emph{universally recognized} for its
eerie connotations and its association with the supernatural\ldots''\\[4pt]
\textbf{Verification:} the injected response is confidently phrased yet
factually wrong; the chosen response remains the best
$\Rightarrow$ \textbf{retained}.
\end{tcolorbox}

\begin{tcolorbox}[enhanced, colback=red!4, colframe=red!50!black,
  colbacktitle=red!50!black, coltitle=white, fonttitle=\bfseries\footnotesize,
  title={Discarded example ($\times$ preference order reversed)},
  boxrule=0.5pt, arc=1mm, left=2mm, right=2mm, top=1.5mm, bottom=1.5mm]
\footnotesize
\textbf{Query:} \emph{I want to argue with you}\\[4pt]
\textbf{Chosen response:} ``Sure, I'd be happy to engage in a respectful and
civil debate with you! What topic do you have in mind\ldots''\\[4pt]
\textbf{Rejected response (beauty-injected):} ``\#\# Debate Guidelines ---
Engaging in a productive and respectful argument is an essential skill. Here
are some \textbf{steps to consider} when you want to argue effectively\ldots''
(a well-structured and genuinely informative guide)\\[4pt]
\textbf{Verification:} the rewriting inadvertently improves the substantive
quality beyond the chosen response $\Rightarrow$ \textbf{discarded}.
\end{tcolorbox}
\caption{\revised{Illustrative examples of the preference verification on the
bias-augmented data: a retained example (top) and a discarded example
(bottom).}}
\label{fig:verif_examples}
\end{figure}

\subsubsection{Bias-aware preference data proportion}
In this part, we would like to investigate how the proportion of bias-aware preference data affects the performance of both generative and discriminative judges. Specifically, for generative judges, we vary the ratio between bias-aware preference pairs and original preference pairs in the training set; for discriminative judges, we vary the number of bias-augmented rejected responses in the contrastive objective.

As shown in Table~\ref{tab:bias_data_ratio}, the proportion of bias-aware supervision exhibits a clear trade-off between bias robustness and general evaluation capability. For generative judges, introducing a small proportion of bias-aware preference data substantially reduces bias sensitivity. Increasing this proportion further yields diminishing returns and gradually degrades performance on general judge benchmarks. Similarly, for discriminative judges, adding bias-augmented rejected responses improves bias resistance but harms general evaluation performance when too many biased rejected responses are used. This suggests that excessive bias-aware rejected responses can dilute the primary quality signal and encourage overfitting to bias-related attributes.

Overall, these results show that bias-aware supervision should be carefully calibrated. While a limited amount of bias-aware data effectively improves robustness to biased evaluations, excessive bias-focused supervision undermines general judging capability by distorting the underlying preference signal. This highlights the importance of controlling the proportion of bias-aware attributes to achieve a balanced trade-off.

\begin{table}[t]
\scriptsize
\setlength{\tabcolsep}{4.7pt}
\begin{tabular}{llllllllll}
\toprule
\multirow{2}{*}{Model} & \multirow{2}{*}{Ratio}
& \multicolumn{3}{c}{JudgeBiasBench}
& RewardBench & JudgeBench & RMB & RM-Bench \\
\cmidrule(lr){3-5} \cmidrule(lr){6-9}
& & $\text{Acc}_{\text{ori}}$ & $\text{Acc}_{\text{inj}}$ & BSR
& Agreement & Agreement & Agreement & Agreement \\
\midrule

\multicolumn{9}{l}{\textbf{Generative judges}} \\
\midrule
\multirow{5}{*}{Qwen2.5-7B}
& 0:4
& 73.1 & 64.9 & 20.7
& \textbf{80.4} & \textbf{53.4} & \textbf{61.6} & \textbf{61.8} \\

& 1:4
& 72.1 & 77.4 & \textbf{10.8}
& 79.7 & 49.7 & 60.7 & 59.3 \\

& 2:4
& \textbf{76.0} & 76.2 & 15.1
& 75.0 & 48.3 & 59.7 & 59.5 \\

& 3:4
& 70.0 & \textbf{78.9} & 11.1
& 72.3 & 51.7 & 57.3 & 58.2 \\

& 4:4
& 71.3 & 78.2 & 11.9
& 75.2 & 48.0 & 53.2 & 56.5 \\

\midrule
\multicolumn{9}{l}{\textbf{Discriminative judges}} \\
\midrule
\multirow{3}{*}{Qwen2.5-7B}
& 0:1
& 74.3 & 58.4 & 30.4
& 87.4 & \textbf{62.3} & \textbf{72.6} & 67.5 \\

& 1:1
& 75.5 & 80.5 & 12.2
& \textbf{90.6} & 58.3 & 71.6 & \textbf{71.3} \\

& 4:1
& \textbf{76.2} & \textbf{81.4} & \textbf{11.3}
& 89.1 & 57.1 & 71.3 & 70.0 \\

\bottomrule
\end{tabular}
\caption{Effect of bias-aware preference data proportion on generative and discriminative judges.
For generative judges, the ratio denotes the proportion of bias-aware preference pairs to original preference pairs in the training set; for discriminative judges, it denotes the ratio between bias-augmented rejected responses and original rejected responses (which both serve as the negatives) in the contrastive objective.}
\label{tab:bias_data_ratio}
\end{table}

\subsubsection{Judge model scaling}

\begin{figure}[t]
    \centering
    \includegraphics[width=0.9\linewidth]{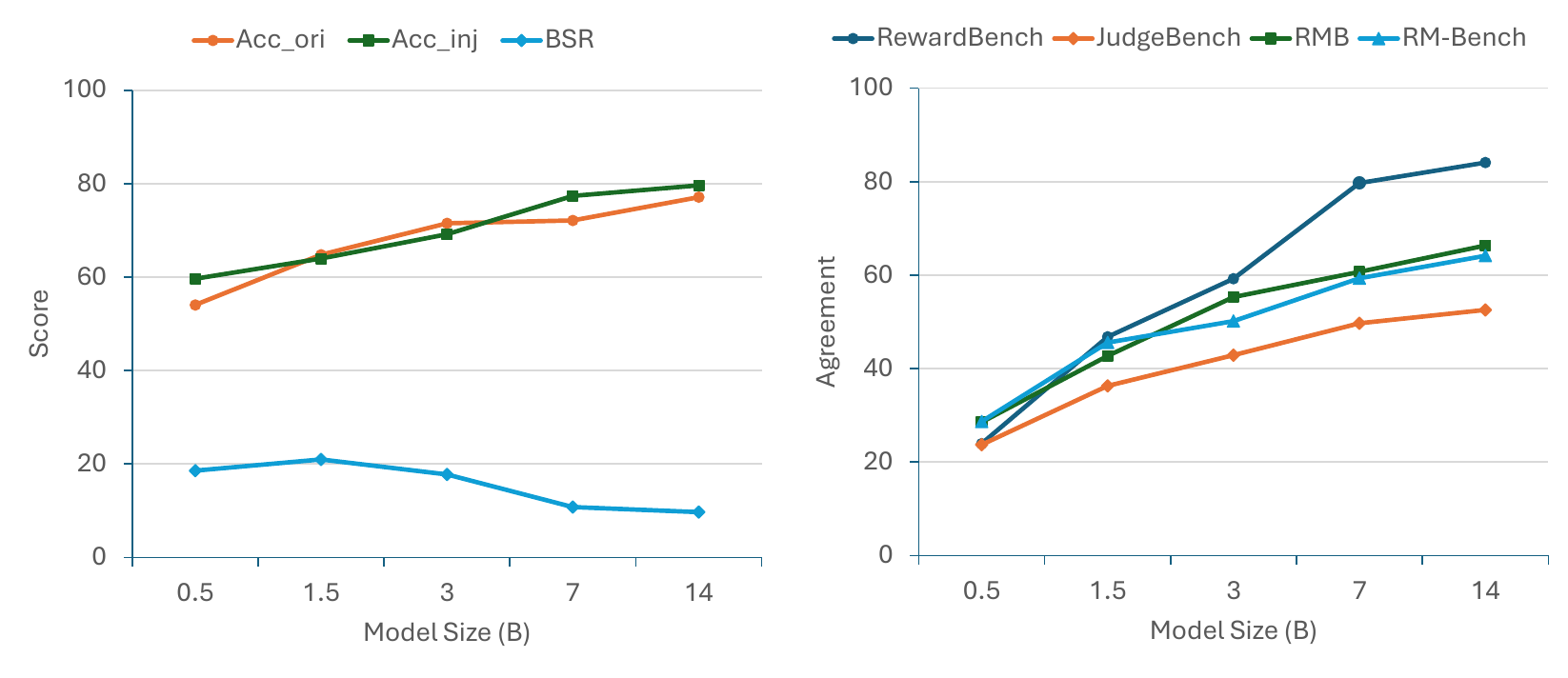}
    \caption{Scaling effects of generative judges.}
    \label{fig:gen_scale}
\end{figure}

\begin{figure}[t]
    \centering
    \includegraphics[width=0.9\linewidth]{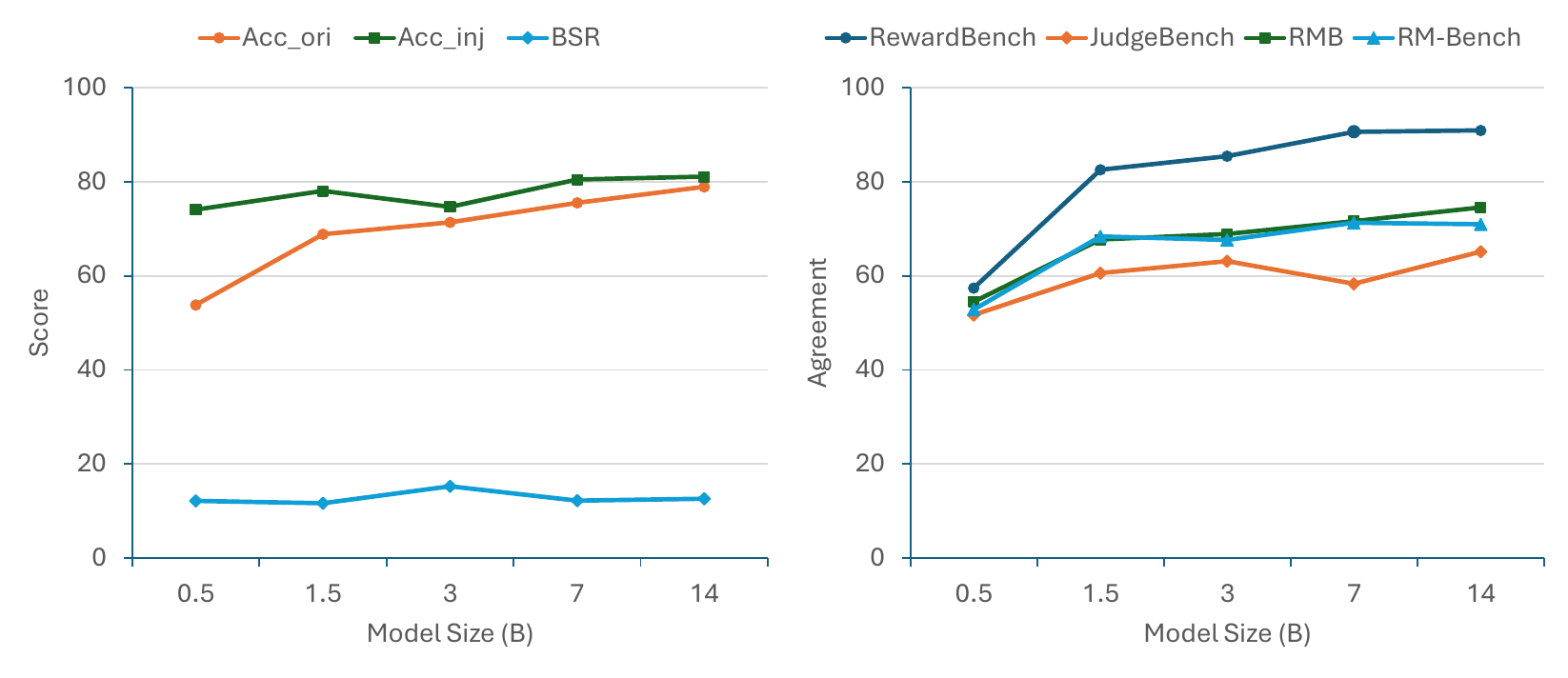}
    \caption{Scaling effects of discriminative judges.}
    \label{fig:dis_scale}
\end{figure}

In this section, we study how judge model scaling affects judgment performance. We use Qwen2.5 models ranging from 0.5B to 14B parameters and apply the same optimization pipeline for both generative and discriminative judges, as described in Section~\ref{sec:bias-mitigation}.

Results on JudgeBiasBench reveal consistent improvements in judgment accuracy as model size increases. For both generative and discriminative judges, $Acc_{ori}$ and $Acc_{inj}$ gradually rise with larger models, indicating that scaling strengthens the ability to identify higher-quality responses even under bias perturbations.

However, BSR exhibits different scaling behaviors for the two judging paradigms. For generative judges, BSR decreases noticeably as model size grows. This suggests that larger generative judges rely less on superficial bias-related cues and are better able to focus on the intrinsic quality of responses. In contrast, discriminative judges maintain consistently low BSR across different scales, with only minor fluctuations. This stability likely stems from the discriminative training objective, which optimizes relative score differences between responses and may encourage the model to focus more on comparative quality signals.

Performance on general judge benchmarks shows a clear scaling trend for both paradigms. Agreement scores on RewardBench, JudgeBench, RMB, and RM-Bench steadily increase with model size, indicating that larger models provide stronger overall evaluation capability. Taken together, these results suggest that model scaling primarily improves judgment accuracy, while bias-aware training maintains stable robustness to bias across different model sizes.

\section{Conclusion}

In this work, we conduct a systematic study of judgment biases in LLM-based judges, which are increasingly used for automated evaluation and reward modeling. To facilitate this analysis, we introduce JudgeBiasBench, a taxonomic benchmark that enables controlled bias injection and systematic measurement of bias effects. Extensive experiments across both generative and discriminative judges reveal that judgment bias remains prevalent and can substantially affect evaluation reliability, even for recent large-scale models. To address this issue, we propose a bias-aware training framework that incorporates bias-related signals during training, mitigating several common bias patterns while preserving strong judgment accuracy. Despite these improvements, our approach mainly addresses bias from the data and training objective perspective, and bias in LLM-based judges remains far from solved. In the future, we will explore more robust model architectures, training paradigms, and evaluation strategies to further improve the fairness and reliability of LLM-based judges.

\section*{CRediT authorship contribution statement}
Hongli Zhou: Writing – original draft, Software, Investigation; Hui Huang: Writing – review and editing, Methodology, Conceptualization; Rui Zhang: Data curation, Validation, Investigation; Kehai Chen: Writing – review and editing; Bing Xu: Writing – review and editing; Conghui Zhu: Supervision; Tiejun Zhao: Supervision; Muyun Yang: Supervision.

\section*{Acknowledgments}
This work was supported by the National Key R\&D Program of China (Grant No. 2025YFC3308800), the National Natural Science Foundation of China (Grant No. 62276077, 62376075, 62376076), the Natural Science Foundation of Heilongjiang Province (Grant No. ZL2025F004), Guangdong Province Basic and Applied Basic Research Fund (Guangdong-Foshan Joint Funds, 2024A1515140026),  the Social Science Foundation of Guangdong Province, China (No. GD24CZY02), and the MOE Key Laboratory of Cognitive Intelligence and Content Security (Grant No. 10120251103, Harbin Institute of Technology).







\bibliographystyle{elsarticle-num}
\bibliography{reference}



\end{document}